\documentclass{article}
\usepackage{arxiv}

\usepackage[utf8]{inputenc} % allow utf-8 input
\usepackage[T1]{fontenc}    % use 8-bit T1 fonts
\usepackage{hyperref}       % hyperlinks
\usepackage{url}            % simple URL typesetting
\usepackage{booktabs}       % professional-quality tables
\usepackage{amsfonts}       % blackboard math symbols
\usepackage{nicefrac}       % compact symbols for 1/2, etc.
\usepackage{microtype}      % microtypography
\usepackage{lipsum}		% Can be removed after putting your text content
\usepackage{natbib}
\usepackage{doi}

\usepackage{amsmath}
\usepackage{amssymb}
\usepackage{graphics}      % for EPS, load graphicx instead 
\graphicspath{{./figures/}}
\usepackage{subcaption}
\usepackage{color}
\usepackage{todonotes}

\usepackage{array}
\usepackage{multirow}
\usepackage{arydshln}
\newcolumntype{L}[1]{>{\raggedright\let\newline\\\arraybackslash\hspace{0pt}}m{#1}}
\newcolumntype{C}[1]{>{\centering\let\newline\\\arraybackslash\hspace{0pt}}m{#1}}
\newcolumntype{R}[1]{>{\raggedleft\let\newline\\\arraybackslash\hspace{0pt}}m{#1}}

\usepackage{setspace}
\usepackage{rotating}

\DeclareMathOperator*{\argmax}{\rm argmax}

\newcommand{\R}{\mathbb{R}}

\newcommand{\abs}[1]{\lvert#1\rvert}				% absolute value
\newcommand{\set}[1]{\left\{#1\right\}}				% set
\newcommand{\vect}[1]{\left[#1\right]}				% vector
\newcommand{\card}[1]{\lvert#1\rvert}				% set size or cardinality

\newcommand{\T}{\mathcal{T}}                        % trace set
\renewcommand{\t}{\tau}                             % trace
\newcommand{\x}{x}                                  % itemset, ie set of features
\newcommand{\f}{f}                               % feature
\newcommand{\F}{\mathcal{F}}                               % feature set

\newcommand{\E}{\Psi}                               % embedding function
\newcommand{\e}{\psi}                               % embedding element
\newcommand{\V}{\mathcal{V}}                        % alphabet

\newcommand{\False}{\mathtt{False}}
\newcommand{\True}{\mathtt{True}}
\newcommand{\OR}[2]{#1\;\mathtt{OR}\;#2}
\newcommand{\AND}[2]{#1\;\mathtt{AND}\;#2}
\newcommand{\UNTIL}[2]{\mathtt{U}(#1,#2)}
\newcommand{\NOT}[1]{\mathtt{NOT}(#1)}
\newcommand{\IMPLIES}[2]{#1\;\Rightarrow\;#2}
\newcommand{\X}[1]{\mathtt{X}(#1)}
\newcommand{\FUT}[1]{\mathtt{F}(#1)}
\newcommand{\GLO}[1]{\mathtt{G}(#1)}
\newcommand{\UNTILINT}[4]{\mathtt{U}[#1,#2](#3,#4)}
\newcommand{\FUTINT}[3]{\mathtt{F}[#1,#2](#3)}
\newcommand{\GLOINT}[3]{\mathtt{G}[#1,#2](#3)}
\newcommand{\UNTILINTSOFT}[5]{\mathtt{U}[#1,#2;#3](#4,#5)}
\newcommand{\GLOINTSOFT}[4]{\mathtt{G}[#1,#2;#3](#4)}
\newcommand{\UNTILSOFT}[3]{\mathtt{U}[#1](#2,#3)}
\newcommand{\GLOSOFT}[2]{\mathtt{G}[#1](#2)}

\newcommand{\Booleans}{\mathbb{B}}

\newcommand{\dkl}{D_{KL}}
\newcommand{\onefunc}{\mathbf{1}}
\newcommand{\Ber}{\mathcal{B}}
\newcommand{\frm}{\varphi}
\newcommand{\Frm}{\Phi}
\newcommand{\Tau}{T}
\newcommand{\Nu}{N}
\newcommand{\policyrate}{\overline{\frm}}
\newcommand{\randomrate}{\widetilde{\frm}}
\newcommand{\satisfies}{\vdash}

\newcommand{\eg}{\textit{e.g.},~}

\newcommand{\ie}{\textit{i.e.},~}

\newcommand{\thetitle}{A Framework for Understanding and Visualizing Strategies of RL Agents}
\newcommand{\thekewywords}{Explainable AI, Competency Awareness, Reinforcement Learning, Strategy Inference, Temporal Logic, Strategy Visualization}
\newcommand{\theauthors}{Pedro Sequeira, Daniel Elenius, Jesse Hostetler, Melinda Gervasio}

\newcommand{\theabstract}{
Recent years have seen significant advances in explainable AI as the need to understand deep learning models has gained importance with the increased emphasis on trust and ethics in AI. Comprehensible models for sequential decision tasks are a particular challenge as they require understanding not only individual predictions but a series of predictions that interact with environmental dynamics. We present a framework for learning comprehensible models of sequential decision tasks in which agent strategies are characterized using temporal logic formulas. Given a set of agent traces, we first cluster the traces using a novel embedding method that %combines discounted feature sums with sequence graph transforms to 
captures frequent action patterns. We then search for logical formulas that explain the agent strategies in the different clusters. %, using the principle of maximum entropy to select the most likely task specifications. 
We evaluate our framework on combat scenarios in StarCraft II (SC2), using traces from a handcrafted expert policy and a trained reinforcement learning agent. %, analyzing performance through both qualitative visualizations and quantitative metrics.
We implemented a feature extractor for SC2 environments that extracts traces as sequences of high-level features describing both the state of the environment and the agent's local behavior from agent replays. 
We further designed a visualization tool depicting the movement of units in the environment that helps understand how different task conditions lead to distinct agent behavior patterns in each trace cluster.
Experimental results show that our framework is capable of separating agent traces into distinct groups of behaviors for which our approach to strategy inference produces consistent, meaningful, and easily understood strategy descriptions.
}

\title{\thetitle}

\author{
\theauthors\\
SRI International\\
333 Ravenswood Ave., Menlo Park, 94025 CA\\
\texttt{\{first\_name.last\_name\}@sri.com}
}

\renewcommand{\headeright}{}
\renewcommand{\undertitle}{}
\hypersetup{
pdftitle={\thetitle},
pdfsubject={\theabstract},
pdfauthor={\theauthors},
pdfkeywords={\thekewywords},
}

\begin{document}

\maketitle

% ABSTRACT
% ------------------------------------------------------------------------
\begin{abstract}
\theabstract
\end{abstract}


% TEASER FIGURE
% ------------------------------------------------------------------------
\begin{figure*}[b]
    \centering
    \includegraphics[width=\textwidth]{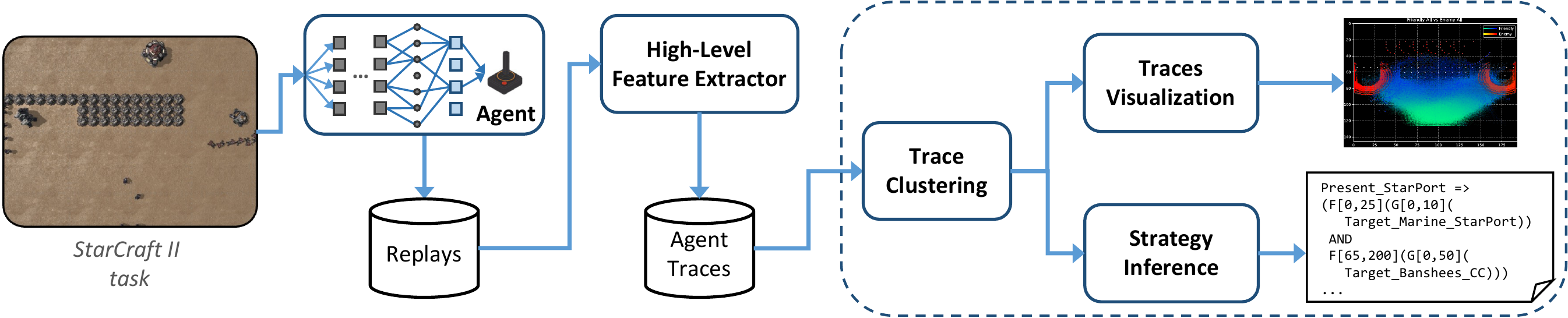}
    \caption{Our framework for discovering and visualizing strategies of RL agents via temporal logic abstraction. Left: the pipeline for recording traces of agent behavior given high-level features. % extracted from StraCraft~II combat scenarios; 
    Right: the framework for extracting strategy descriptions via trace clustering and temporal logic inference, and the traces visualization tool.}
    % \Description{Our pipeline for discovering strategies of RL agents in StarCraft II via temporal logic abstraction. We start by producing a set of agent behavior replays from the StarCraft II game engine, which we then convert to a set of agent traces by applying a high-level feature extractor. The traces are fed to a clustering algorithm that partitions the traces by capturing different behavioral patterns, and produces visualizations for each cluster. Each trace cluster is finally fed to the strategy inference component that extracts a set of strategy descriptions in the form of temporal logic formulas.}%
	\label{Fig:Framework}
\end{figure*}
% ------------------------------------------------------------------------

% INTRODUCTION
% =========================================
\section{Introduction}%
\label{Sec:Intro}

% motivation
Rapid advances in deep learning have resulted in phenomenal successes for artificial intelligence (AI) over the last several years. Machine-learning-based AI systems have achieved superhuman performance on a number of image classification tasks and complex games and now see routine use in customer service, machine translation, robotics, recommender systems, and more. Deep reinforcement learning (RL), in particular, is behind the success of a number of game-playing systems that have mastered complex games such as Go \citep{silver2018rl}, StarCraft \citep{vinyals2019alphastar}, Texas Hold'Em \citep{heinrich2016deep}, and Dota 2 \citep{berner2019dota2}. However, RL models are black boxes: it is hard to understand what an agent has learned, to determine the conditions that affect its behavior, and to assess whether it knows what to do in different situations. Yet, understanding the strategies employed by autonomous agents is an essential component of trust, which is critical to the acceptance of autonomous intelligent systems. 

Explainable AI seeks to address this problem by developing techniques that enable machine learning systems to explain their decisions or actions to human users. In explainable RL, this has often translated to the development of methods for explaining policy network decisions or identifying significant decision points. However, user studies have shown the value of richer explanations---\eg users prefer explanations about policies to explanations about single actions \citep{vanderwaa2018contrastive}, and contrastive explanations enable better task prediction \citep{madumal2020causallens}. Here, we present an approach to explaining and visualizing the policies that govern an agent's actions by abstracting traces of agent behavior into meaningful strategies in a form that is understandable even to non-technical end users.

% overview of framework for extracting strategy rules (clustering + inference) given behavior traces
Our framework for extracting meaningful descriptions of agent strategy from traces of agent behavior is depicted on the right half of Fig.~\ref{Fig:Framework}. The input is a set of traces of agent behavior on a task, where a trace corresponds to a variable-length sequence of observations, each a fixed-length vector of categorical features. Our approach involves first partitioning the traces into a set of clusters, each capturing distinct aspects of the agent's behavior on that task. 
%To that end, we propose 
We employ a novel embedding method that converts each trace to a numeric vector that combines discounted feature sums to represent condition features and sequence graph transforms \citep{ranjan2016sequence} to capture frequent patterns over action features. We then cluster the resulting embeddings for each trace using hierarchical agglomerative clustering (HAC). Finally, we employ an information-theoretic approach to strategy inference to capture strategies as soft metric temporal logic formulas, using a learning-from-demonstration approach based on the principle of maximum entropy. The result is a set of logical formulas that describe the strategies employed by the agent in each cluster.

% experiments in SC2, expert + RL agent, etc.
We evaluated our framework within the StarCraft~II (SC2) Learning Environment \citep{vinyals2017starcraft}, as depicted on the left side of Fig.~\ref{Fig:Framework}, by designing different force-on-force combat scenarios. Overall, our scenarios favor different strategies for achieving the primary objective of capturing the enemy command center, depending on force matchups, existing secondary objectives, the presence of barriers, and the random arrival of enemy reinforcements. We performed strategy inference on traces collected from both a handcrafted expert agent using a deterministic ``smart'' policy, and a deep RL agent trained on the combat task. 
To allow learning interpretable strategy descriptions, we implemented a high-level feature extractor for SC2 environments that describes the elements of the task (the types and properties of units, amount and size of forces, etc.) and abstracts the agent's local behavior (the movement of groups of units relative to the opponent, the game actions assigned to each unit type, etc.). 
Further, to assist in the interpretation of agent behavior in the discovered clusters, we designed a tool for visualizing the spatio-temporal dynamics of the friendly and enemy units in the traces of each cluster. The results of our experiments show that our clustering approach is able to partition traces into clusters that exhibit distinct strategies and that the logical formulas extracted by our strategy inference component effectively capture important aspects of the behaviors in the clusters.

The primary contribution of this paper is a model-agnostic framework for extracting meaningful descriptions of agent strategy from traces of agent behavior. As highlighted in Fig. 1, the framework is composed of:
\begin{itemize}
  \item A high-level feature extractor for describing the state of the environment and the agent's local behavior that works with any type of SC2 scenario.
  % that works with any type of SC2 scenario that allows describing both the state of the environment and the agent's local behavior.
  \item A novel embedding method for representing behavior traces that combines discounted binary feature sums %, capturing the distribution of condition features over time, 
  with Sequence Graph Transform (SGT), which we extend to operate over sequences of symbols.
  %. %, capturing temporal relationships between agent actions. 
  % To allow clustering traces using multiple features, we also extend SGT from sequences of (single) symbols to sequences of itemsets, \ie sequences of sets of symbols.
  \item A simple but effective tool to visualize aggregate spatio-temporal patterns in a set of traces that helps understand how different task conditions lead to distinct agent behaviors.
  \item An elegant, composable, information-theoretic approach to inferring strategies in the form of soft metric temporal logic formulas.
\end{itemize}

The paper is organized as follows: Section~\ref{Sec:Framework} details the components of our framework for discovering and visualizing agent strategies from behavior traces, Section~\ref{Sec:Experiments} outlines our experiments in the StarCraft II domain used to evaluate our approach, and in Section~\ref{Sec:Results} we show and analyze the results of the experiments. In Section~\ref{Sec:Related} we discuss related work and in Section~\ref{Sec:Conclusions} we outline the main conclusions and present current and future research directions.

% STRATEGY DISCOVERY FRAMEWORK
% =========================================
\section{Discovering Agent Strategies}%
\label{Sec:Framework}

We now detail the agent traces that are input to our system, followed by the two main components of our approach: Trace Clustering and Strategy Inference.

\subsection{Agent Traces}
\label{Subsec:Traces}
% definition of trace as sequence of itemsets
The input to our framework is a set of agent traces for which we want to discover meaningful, easily understood formulas that summarize the agent's strategies in the task of interest. Formally, let $\F=\set{\f_1,\ldots,\f_m}$ be a finite set of \emph{features}, each corresponding to some qualitative aspect of the environment or the agent's behavior. An \emph{observation}, denoted by $o: \F \to \Booleans$, where $\Booleans = \set{0,1}$, assigns a Boolean value to each feature, and represents the state of the world at some discrete timestep. A \emph{trace}, denoted by $\tau = (o_1,\ldots,o_n)$, is a finite sequence of observation vectors corresponding to a single episode of the agent's interaction with the environment. We assume that traces can be of varying length, but that the feature space $\F$ is fixed. The value of feature $\f$ at timestep $t$ in trace $\t$ can be written as $\t_t(\f)$. Whenever the trace is implicit, we write $\f^t$ to denote the value of feature $f$ at time $t$. In addition, we can represent a trace $\t$ as a matrix of size $\card{\t} \times \card{\F}$.

% ------------------------------------------------------------------------
\subsection{Trace Clustering}%
\label{Subsec:Clustering}
% clustering overview, embedding definition
%As seen in Fig.~\ref{Fig:Framework}, we first cluster the traces before performing strategy inference. 
The goal of Trace Clustering is to partition the traces based on distinctive aspects of the agent's behavior under different environmental conditions that lead to different ways of solving the task. To apply standard clustering algorithms over traces, we first need to define a common representation for each trace such that traces can be compared to each other. Since traces are of different lengths, one option is to learn an embedding representation for each trace. Formally this corresponds to defining a function $\E:\t\to\R^n$ mapping a trace to a numeric vector. 
%Here we
We consider two types of embeddings for representing agent traces: \emph{feature counts} and \emph{Sequence Graph Transform} (SGT).

% ________________________________________________________________________
%\subsubsection{Feature Counts Embedding}
\subsubsection{Trace Embeddings}
% feature sums definition
\emph{Feature counts} is a common method for characterizing agent trajectories in the context of Markov Decision Processes (MDPs). Formally, the feature count embedding function corresponds to $\E_{fc}(\t)=\vect{\e_0,\e_1,\ldots,\e_n}$ where each embedding feature $\e_i=\sum_{t=0}^{\card{\t}}{\gamma^t\f_i^t}:i=0,\ldots,n=\card{\F}$ computes the discounted sum of features $\f_i$ in trace $\t$, with discount factor $\gamma\in[0,1]$. The discounted version enables us to distinguish between the appearance of features earlier in an episode versus at the end.

% ________________________________________________________________________
% \subsubsection{SGT Embedding}

% problem with feature counts, SGT intro
A potential problem with feature counts is that only the total (discounted) frequencies of features are considered, thus disregarding potential temporal dependencies between features. In our work, we are particularly interested in the relationships between action-related features such that we can cluster traces capturing different action patterns of the agent. To address this, we use \emph{Sequence Graph Transform} (SGT) \citep{ranjan2016sequence}, a recently proposed approach for characterizing sequences of symbols. SGT computes an embedding that captures short- and long-term forward dependencies between each pair of symbols appearing in a set of sequences of interest. %Some advantages of SGT are that it supports traces of different lengths, captures both short- and long-term feature dependencies (temporal patterns), and the resulting embedding can be represented as a graph where nodes correspond to symbols and edges to the temporal dependency between symbols in a sequence.

% original SGT definition
We refer to the set of all symbols as the \emph{alphabet}, and denote it by $\V$. In our setting, the symbols correspond to the truth values assigned to each Boolean feature \ie $\V=\bigcup\limits_{\f\in\F}{\set{\f=0,\f=1}}$. %Following the notation introduced above, 
Given a trace $\t\in\T$, for each symbol pair $(u,v)\in\V\times\V$, SGT computes an \emph{association feature} between $u$ and $v$, denoted by $\e_{uv}$, as follows:%
\footnote{Here we use the \emph{length-sensitive} version of the SGT association feature \citep{ranjan2016sequence}, in which sequences of similar symbol patterns but of considerably different lengths are \emph{not} scored as being close to each other in embedding space.}
\begin{displaymath}
    \e_{uv}(\t)=\frac{\sum_{\forall{(l,m)\in\Lambda_{uv}}}{\rho(l,m)}}{\card{\Lambda_{uv}}},
\end{displaymath}
where $\Lambda_{uv}$ comprises the timestep indices $l,m$ where symbols $u,v$ are satisfied in trace $\t$, respectively. $\rho(l,m)=\epsilon^{-\kappa\abs{m-l}}$ measures the ``temporal effect'' of events $m$ and $l$, and $\kappa$ is a hyper-parameter influencing the long-term dependency effect size. 

% extension to support sequences of itemsets
The original formulation in \citep{ranjan2016sequence} considers only sequences of symbols whereas our agent traces contain sequences of sets of features/symbols as defined above. Therefore, we extended SGT to consider associations between all pairs of features $\f\in\F$, including between the truth values of each feature. This results in the following modified version of the timestep indices selection function: $\Lambda_{uv}=\set{(l,m):u\in\x_l,v\in\x_m,l<m,l,m\in\set{0,\ldots,\card{\t}}}$. The SGT embedding function corresponds to $\E_{sgt}(\t)=\vect{\e_{uv}(\t):(u,v)\in\V\times\V}$, \ie the concatenation of the association features for all pairs of symbols in the alphabet.

% ________________________________________________________________________
\subsubsection{Clustering Traces}

% trace embedding via FC + ST combination
%Our objective in
The objective of Trace Clustering is to partition traces to capture not only distinct agent behaviors but also the different effects of those behaviors in the environment. SGT lets us capture such patterns but, in our preliminary experiments, we found no advantage to computing SGT embeddings for all available features. In particular, using feature counts for the condition features and computing SGT for the action-related features %seems to result 
resulted in better separated, more meaningful clusters that captured distinct strategies while also being less computationally expensive. As such, in our experiments we use the following embedding function:%
\begin{equation}\label{Eq:Embed}
    \E(\t)=\vect{\E_{sgt}(\t_{act}),\E_{fc}(\t_{task})},
\end{equation}
where $\t_{act}$ corresponds to the sequence of action-related features in trace $\t$ and $\t_{task}$ to the sequence of condition features. Further, we denote by $\F_{act}$ the space of action-related features and by $\F_{task}$ the space of condition features.

% normalization, filtering
After computing the embeddings for each trace $\t\in\T$ using Eq.~\ref{Eq:Embed}, we normalize them to the $[0,1]$ interval via min-max scaling, which is needed since we are combining different embedding functions. We then filter the embedding features by removing those whose values are constant across {\it all} traces, which both helps to reduce the size of the embedding representation for clustering and leads to better interpretability by having a smaller pool of features.

% clustering via HAC, internal eval
Finally, we cluster the traces using Hierarchical Agglomerative Clustering (HAC) \citep{kaufman1990agglomerative}. %HAC is a bottom-up clustering algorithm that starts by assigning a different cluster to each datapoint and then iteratively combines pairs of clusters according to some linkage criterion and distance function  until a single cluster, containing all datapoints, is reached. 
HAC has the advantage of not requiring the number of clusters to be specified \textit{a priori}. To select the number of clusters, we compute the \emph{Cali\'{n}ski-Harabasz score} \citep{calinski1974dendrite}, an internal clustering evaluation metric that, given a data partition (number of clusters), informs how differentiated the datapoints from different clusters are compared to how dispersed the datapoints are within in each cluster. %computes the ratio of the sum of between-clusters dispersion and the sum of within-cluster dispersion of all clusters, where the dispersion is defined as the sum of the squared pairwise distances.
%\emph{silhouette coefficient} \citep{rousseeuw1987silhouettes}. This is an internal evaluation metric that assesses, for each cluster, how close the traces within the cluster are to each other compared to how distant each trace is to traces in the next nearest cluster. 
We then choose the number of clusters for which this score is highest, corresponding to the partitioning with %better-defined
the most well-defined clusters.

% ------------------------------------------------------------------------
\subsection{Temporal Logic}\label{Sec:logic}

The next step in our approach is Strategy Inference, the goal of which is to find strategy descriptions in the form of temporal logic formulas. Temporal logic is a family of logic languages that allows expressing properties over temporal sequences of events. Using temporal logic to express agent strategies has a number of advantages~\citep{NEURIPS2018_taskspec}: unlike an RL agent's reward function or policy network, temporal logic formulas can be rendered into succinct natural language descriptions, transferred to different environments, and composed into larger strategies.
%
%In this work, we 
Here we focus on the \emph{future} variant of the widely used Linear Temporal Logic (LTL)~\citep{Pnueli1977TheTL}, which  %LTL allows for both {\it past} and {\it future} variants. In past LTL, we can specify what has happened {\it up until now}, whereas in future (normal) LTL, we can specify {\it what will happen in the future}. Past LTL is useful for online monitoring~\citep{ulus_ltl}, but future LTL to be more natural for expressing strategies.
%In addition, future LTL gives us the ability to use 
allows specifying strategy formulas for \emph{prediction}, \eg to say that an agent will eventually do action $A$ given current conditions $C$.

\subsubsection{Linear Temporal Logic (LTL)}
We now formally define the basic future LTL logic. First, the boolean features $\f \in \F$ are used as atomic propositions. LTL formulas are then built using the following grammar:
\begin{equation*}
    \frm = \True\; | \;\False\; |\;\f\; | \;\NOT{\frm}\; | \;\OR{\frm_1}{\frm_2}\; | \;\AND{\frm_1}{\frm_2}\; | \;\IMPLIES{\frm_1}{\frm_2}\; | \;\X{\frm}\; | \;\UNTIL{\frm_1}{\frm_2}\; | \;\FUT{\frm}\; | \;\GLO{\frm}\;
\end{equation*}
We then define a satisfaction relation $(\tau,t) \satisfies \frm$ that indicates whether trace $\tau$ satisfies formula $\frm$ at time $t$ according to:
\begin{alignat*}{1}
    % (\tau, t) \satisfies \True                     &\\
    (\tau, t) \satisfies \f                        \iff & \t_t(\f) = 1 \\
    % (\tau, t) \satisfies \NOT{\frm}                \iff & (\tau, t) \notsatisfies \frm \\
    % (\tau, t) \satisfies \OR{\frm_1}{\frm_2}       \iff & (\tau, t) \satisfies \frm_1 \;\textrm{or}\; (\tau, t) \satisfies \frm_2 \\
    (\tau, t) \satisfies \X{\frm}                  \iff & (\tau, t+1) \satisfies \frm\\
    (\tau, t) \satisfies \UNTIL{\frm_1}{\frm_2}    \iff & \exists_{t' \geq t} (\tau, t') \satisfies \frm_2 \;\textrm{and}\; \forall_{t''\in [t,t')} (\tau, t'') \satisfies \frm_1\\
% \end{alignat*}
%
% We are often interested in whether a formula is satisfied by a trace as a whole, \ie for $t=0$. For this we use the short-hand $\tau \satisfies \frm$.
% We can now also define the following derived operators:
%
% \begin{alignat*}{1}
    % \False &\equiv \NOT{\True}\\
    % \AND{\frm_1}{\frm_2} &\equiv \NOT{\OR{\NOT{\frm_1}}{\NOT{\frm_2}}}\\
    % \IMPLIES{\frm_1}{\frm_2} &\equiv \OR{\NOT{\frm_1}}{\frm_2}\\
    (\tau, t) \satisfies \FUT{\frm}                 \iff & \UNTIL{\True}{\frm}\\
    (\tau, t) \satisfies \GLO{\frm}                 \iff & \NOT{\UNTIL{\True}{\NOT{\frm}}},\\
\end{alignat*}
where the \texttt{NOT}, \texttt{OR}, \texttt{AND} and $\Rightarrow$ (implies) operators compute the corresponding propositional logic operation. We call $\mathtt{X}$, $\mathtt{U}$, $\mathtt{F}$, and $\mathtt{G}$ {\it temporal operators}, since their truth value depends on the truth values of their sub-formulas at {\it other} time steps. Intuitively, they can be understood as follows: $\X{\frm}$: $\frm$ is true in the next time step; $\UNTIL{\frm_1}{\frm_2}$: $\frm_1$ is true for all time steps \emph{until} $\frm_2$ is true; $\FUT{\frm}$: $\frm$ is true in some \emph{future} time step; $\GLO{\frm}$: $\frm$ is \emph{``globally''} true, \ie for all time steps. %The remaining operators are the familiar ones from propositional logic.
Further, we use the short-hand $\tau \satisfies \frm$ to denote whether a formula is satisfied by a trace as a whole, \ie for $t=0$.

\subsubsection{Soft Metric Temporal Logic (SMTL)}
% \subsubsection{Metric Temporal Logic (MTL)}
In addition to the above operators, it is often useful to introduce temporal operators that operate over a %limited 
subset of the timeline. 
%This bring us to 
To accommodate this, we use Metric Temporal Logic (MTL)~\citep{Koymans2005SpecifyingRP}, which adds the following operator: % $\UNTILINT{a}{b}{\frm_1}{\frm_2}$, with the semantics:
%
% \begin{alignat*}{1}
\begin{displaymath}
    (\tau, t) \satisfies \UNTILINT{a}{b}{\frm_1}{\frm_2} \iff \exists_{t' \in [t+a, t+b]} (\tau, t') \satisfies \frm_2 \;\textrm{and}\; \forall_{t'' \in [t,t')} (\tau, t'') \satisfies \frm_1,
\end{displaymath}
% \end{alignat*}
%
%\ie 
where $a$, $b$ define a time interval after $t$ in which formula $\frm_2$ needs to be satisfied. %and we can once again define the derived operators
%
% \begin{alignat*}{1}
    % &\FUTINT{a}{b}{\frm} \equiv \UNTILINT{a}{b}{\True}{\frm}\\
    % &\GLOINT{a}{b}{\frm} \equiv \NOT{\FUTINT{a}{b}{\NOT{\frm}}}\\
% \end{alignat*}
Similarly, we define operators $\FUTINT{a}{b}{\frm}$ and $\GLOINT{a}{b}{\frm}$ as the MTL versions of $\FUT{\frm}$ and $\GLO{\frm}$ defined above, respectively.

% \subsubsection{Soft Metric Temporal Logic (SMTL)}
During preliminary evaluation of temporal logic formulas over traces of RL agents, we noticed that, due to the inherent stochasticity of agent behavior, some formulas were hard to be satisfied in their entirety. % in the behavior of RL agents.
For example, even if an agent performed an action $a$ 95\% of the time, a formula such as $\GLO{a}$ will not be satisfied, as it requires that action $a$ be true in the traces for \emph{every} time step. To accommodate such situations, we introduce a \emph{soft} version of Metric Temporal Logic (SMTL). 
%First, let us denote by 
Let $\kappa(\tau,\frm,a,b) =   |\{t \in [a,b] | (\tau,t) \satisfies \frm\}|$ denote the \emph{satisfaction count} and 
%by 
$\nu(\tau,\frm,a,b) = \frac{\kappa(\tau,\frm,a,b)}{b-a+1}$, the \emph{satisfaction rate}. %First, let us define {\it satisfaction count}, denoted by $\kappa$, and {\it satisfaction rate}, denoted by $\nu$, using:
%
% \begin{alignat*}{1}
    % \kappa(\tau,\frm,a,b) = & \; |\{t \in [a,b]\; | \; (\tau,t) \satisfies \frm\}| \\
    % \nu(\tau,\frm,a,b) = & \; \dfrac{\kappa(\tau,\frm,a,b)}{b-a+1}
% \end{alignat*}
%
We then define the following {\it soft temporal operator}:
%
% \begin{alignat*}{1}
\begin{displaymath}
    % (\tau, t) \satisfies \GLOINTSOFT{a}{b}{r}{\frm} \iff& \nu(\tau,\frm,t+a,t+b) \geq r\\
    (\tau, t) \satisfies \UNTILINTSOFT{a}{b}{r}{\frm_1}{\frm_2} \iff \exists_{t' \in [t+a, t+b]} (\tau, t') \satisfies \frm_2 \;\textrm{and}\; \nu(\tau,\frm_1,t,t'-1) \geq r,
\end{displaymath}
% \end{alignat*}
%
that can be understood as $\frm_1$ being true at least $r$ of the time until $\frm_2$ is satisfied. Similarly, we define $\GLOINTSOFT{a}{b}{r}{\frm}$ as the SMTL version of $\GLOINT{a}{b}{\frm}$, which can be interpreted as $\frm$ being true at least $r$ of the time.%
\footnote{We note that there is no soft version of $\mathtt{F}$, because $\mathtt{F}$ is not required to hold over an interval.}
Further, we define $\GLOSOFT{r}{\frm}$ and $\UNTILSOFT{r}{\frm_1}{\frm_2}$, corresponding to the soft versions of \texttt{G} and \texttt{U} without time interval constraints, respectively. 
%sometimes prefer the following soft operators without intervals:
%
% \begin{alignat*}{1}
    % (\tau, t) \satisfies \GLOSOFT{r}{\frm} \iff& \nu(\tau,\frm,t,|\tau|) \geq r\\
    % (\tau, t) \satisfies \UNTILSOFT{r}{\frm_1}{\frm_2} \iff& \exists t' \in [t, |\tau|].\; (\tau, t') \satisfies \frm_2 \;and\\
                                                            %   &\nu(\tau,\frm_1,t,t'-1) \geq r
% \end{alignat*}
%
% (Note that there is no soft version of $\mathtt{F}$, because $\mathtt{F}$ is not required to hold over an interval). These operators can be understood as follows: $\GLOSOFT{r}{\frm}$: $\frm$ is true at least $r$ of the time; $\UNTILSOFT{r}{\frm_1}{\frm_2}$: $\frm_1$ is true at least $r$ of the time until $\frm_2$ is true.
 
\subsubsection{Evaluating SMTL Satisfaction}

%It is critical to our Strategy Inference algorithm that we can efficiently evaluate the satisfaction of SMTL formulas, $\tau \satisfies \frm$. Fortunately, an efficient evaluation algorithm exists. While the full details are beyond the scope of this paper, we briefly describe the approach here. The core idea of the approach comes from \citep{ulus_ltl}, with changes to account for a) efficiency, b) our added operators, and c) the fact that we are using a future temporal logic, as opposed to the past LTL in \citep{ulus_ltl}.
To efficiently evaluate the satisfaction of SMTL formulas we follow the approach of \cite{ulus_ltl}, with changes to account for: a) efficiency, b) our added operators, and c) the use of future temporal logic.
The approach is based on the observation that the value of any formula $\frm$ in a given time step $t$ only depends on the truth-value of $\frm$ at time steps ${t+1 \ldots |\tau|}$, and of its sub-formulas at time steps ${t \ldots |\tau|}$.%
\footnote{In the original algorithm in \citep{ulus_ltl}, the dependency is only on the {\it next} time step, but our extension with soft temporal operators involves looking further away in time; this does not increase the computational complexity, but it does have a memory cost.}
Therefore, we evaluate formulas starting from the last time step and then proceed backwards until reaching %in time towards
the first time step, starting with atomic sub-formulas and proceeding up towards the complete top-level formulas. The complexity of computing $\tau \satisfies \frm$ is $\mathcal{O}(|\tau| \cdot c)$ %$\mathcal{O}(|\tau| \cdot c)$ where $c$ is the formula complexity, i.e., the average number of sub-formulas per formula. The complexity $c$ is typically a small constant number, so we can ignore it here. Evaluating 
and evaluating $N$ formulas over $M$ traces with average length $K$ thus has (amortized) complexity $\mathcal{O}(N \cdot M \cdot K)$.

% \subsection{Condition Discovery}

% Having found clusters of distinct agent behaviors (see Section \ref{Subsec:Clustering}), we want to characterize the
% initial conditions in each cluster as a logical formula. These initial condition formulas are then used as conditions
% for any strategies discovered for that cluster. The initial condition formulas should be {\it mutually exclusive}
% and {\it exhaustive}, so that we can treat them as a decision procedure for the agent's choice of behavior.
% To achieve this, we first learn a decision tree based on the initial conditions of the traces in each cluster, which tries
% to determine the cluster based on the conditions. We use the well-known C4.5 algorithm~\citep{Quinlan1993}.
% The next step is to convert the decision tree to a set of logical formulas (one per cluster). For each cluster, we extract the 
% sub-tree of the learned decision tree whose leaves (i.e., decisions) match that cluster. Then, we convert the sub-tree
% into a logical formula, by traversing it top-down and introducing an $\mathtt{OR}$ node for any branching of the tree, and an $\mathtt{AND}$ node for any step down a branch. The leaves correspond to atomic formulas, or negated atomic formulas.

\subsection{Strategy Discovery}%
\label{Sec:StrategyDiscovery}

\citet{NEURIPS2018_taskspec} describe an approach to inferring a task specification % (what we call a \emph{strategy}) 
from a set of demonstrations. Specifically, the goal is to find
\begin{equation}\label{taskspec_objective}
    {\frm}^* \in \argmax_{\frm \in \Frm}Pr(\frm|M,X),
\end{equation}
where $Pr(\frm|M,X)$ denotes the probability that the teacher demonstrated the task specification $\frm$ given the observed traces $X$, the dynamics $M$, and the candidate task specifications $\frm$. To solve~\ref{taskspec_objective}, we first define $\Nu(\Tau,\frm) = \frac{|\{\tau \in \Tau\; | \; \tau \satisfies \frm\}|}{|\Tau|}$ 
%corresponding to 
as the satisfaction rate over a set of traces $\Tau$ and a formula $\frm$.
%
% \begin{equation*}
    % \Nu(\Tau,\frm) = \dfrac{|\{\tau \in \Tau\; | \; \tau \satisfies \frm\}|}{|\Tau|}
% \end{equation*}
%
% (The computation of $\tau \satisfies \frm$ was discussed in Section \ref{logic}).
We then collect a set of traces $\overline{\Tau}$ from the agent whose policy we are analyzing, and a second set $\widetilde{\Tau}$ from an agent that takes random actions in the same environment, and use the shorthand $\policyrate \doteq \Nu(\overline{\Tau},\frm)$, $\randomrate \doteq \Nu(\widetilde{\Tau},\frm)$. Following \cite{NEURIPS2018_taskspec}, we can then, under some assumptions, compute Equation \ref{taskspec_objective} as
\begin{equation}\label{taskspec_objective_2}
    {\frm}^* \in \argmax_{\frm \in \Frm}\left\{ \onefunc[\policyrate \geq \randomrate] \cdot \dkl\left(\Ber(\policyrate) \parallel \Ber(\randomrate)\right) \right\},
\end{equation}
where $\mathbf{1}$ is the indicator function, $D_{KL}$ is the Kullback-Leibler (KL) divergence, and $\mathcal{B}(x)$ denotes a Bernoulli distribution with parameter $x$. %Expanding definitions, we get the following closed-form formula:
%
% \begin{equation}%\label{taskspec_objective_2}
    % {\frm}^* \in \argmax_{\frm \in \Frm} 
    % \left\{
    % \mathbf{1}[\policyrate \geq \randomrate] 
    % \left(\policyrate \log\frac{\policyrate}{\randomrate} + (1-\policyrate) \log\frac{1-\policyrate}{1-\randomrate} \right) \right\}
% \end{equation}
%
In other words, our goal is to maximize the KL divergence, with the added requirement that the policy agent's satisfaction rate be larger than the random agent's. We additionally clip the two satisfaction rates to $[\epsilon, 1-\epsilon]$ for some very small number $\epsilon$, to ensure mathematical stability.

% In the SC2 setting, we do not have an explicit teacher showing us demonstrations of a task. However, the clusters we discover through Trace Clustering (see Sec.~\ref{Subsec:Clustering}) are analogous: we can treat each cluster as a distinct set of ``demonstrations'' and we can pair each of the policy agent clusters with the same (un-clustered) set of random traces.
In our experiments, after clustering the agent traces following the procedure in Sec.~\ref{Subsec:Clustering}, we treat each cluster as a set of ``demonstrations'' and pair each cluster with the same (un-clustered) set of traces produced by a random agent. 
To summarize, given a set of candidate %strategy 
formulas $\frm$, and traces for the policy agent and the random agent, we have a method for finding which %strategies 
formulas are the most likely to be "demonstrated" to us by the agent. 

% \subsection{Generation of Candidate Strategies}%
% \label{Sec:StrategyTemplates}

%As mentioned in the previous section, we need to generate temporal logic formulas to serve as candidate strategies to evaluate. %We do not generate arbitrary formulas in a fully random way, because most formulas would not be meaningfully interpreted as strategies. Instead, we 
The remaining issue is how to find the candidate formulas $\Frm$ that best describe the agent's behavior. We observe that there are three main components of an agent's behavior %a strategy 
that we wish to capture in a strategy:
\begin{itemize}
    \item \textbf{Condition}. The situation under which the agent employs the strategy.
    \item \textbf{Action}. The action the agent performs in the strategy.
    \item \textbf{Goal}. The goal the agent is trying to accomplish by employing the strategy.
\end{itemize}
Further, in our work, we distinguish between \emph{strategy}, which describes the overarching objective that the agent wants to achieve, from \emph{tactics}, which correspond to concrete actions the agent needs to execute given certain conditions to achieve some goal. One option is to try to find a combination of tactics denoting an agent strategy directly from the traces. However, in preliminary experiments we determined this to be infeasible due to the combinatorial explosion of possible strategies. Instead, our approach is to first find good individual tactics given the following templates:
% To generate candidate strategies, we use {\it strategy templates}. 
%
% As such, in this paper we use the following strategy templates:
%
\begin{alignat*}{1}
    % \emph{goal}:                  &\;\FUT{G}\\
    % \emph{action}:                &\;\FUT{\GLOSOFT{r}{A}}\\
    \emph{condition-action}:      &\;\FUT{\AND{C}{\X{\GLOINTSOFT{0}{d}{r}{A}}}}\\
    \emph{action-goal}:           &\;\FUT{\UNTILINTSOFT{1}{1000}{r}{\AND{A}{\NOT{G}}}{G}}\\
    % \emph{condition-action-goal}: &\;\FUT{\AND{\AND{C}{\NOT{G}}}{\UNTILINTSOFT{1}{1000}{r}{\AND{A}{\NOT{G}}}{G}}}
\end{alignat*}
where $d$ and $r$ are SMTL parameters as defined above and $C$, $G$, and $A$ are respectively the condition, goal and action feature parameters of the formulas, that we have to fill in and evaluate during %strategy 
the inference process.

We then define a strategy for an agent as the combination of the best tactics discovered for the agent given its traces. In particular, we are interested in identifying tactics where some feature $f$ denotes an important situation used by the agent both as the goal of an action, \ie corresponding to an \emph{action-goal formula} where $G=f$, and also as a condition for an action, \ie corresponding to a \emph{condition-action} formula where $C=f$. We score tactics according to Eq.~\ref{taskspec_objective_2} which aims at differentiating the agent's behavior from that of a random agent. Together, the templates allow us to search for likely actions, goals, conditions, and combinations of these factors, to describe agent behavior.
\section{Experiments}%
\label{Sec:Experiments}

This section details the experiments we conducted to validate our strategy discovery framework. We begin by describing our StarCraft~II (SC2) combat scenario%
\footnote{We chose SC2 because of its widespread use, ease of scenario editing, and availability of supporting tools but note that our approach is domain- and algorithm-agnostic. We use the Python component available at: \url{https://github.com/deepmind/pysc2}.}
and the agents we used in our experiments as well as the SC2 high-level feature extractor, before describing the experiments themselves.

% ------------------------------------------------------------------------
\subsection{StarCraft~II Task}

% ......................................................................................
\begin{figure}
    \centering
    \includegraphics[width=0.6\textwidth]{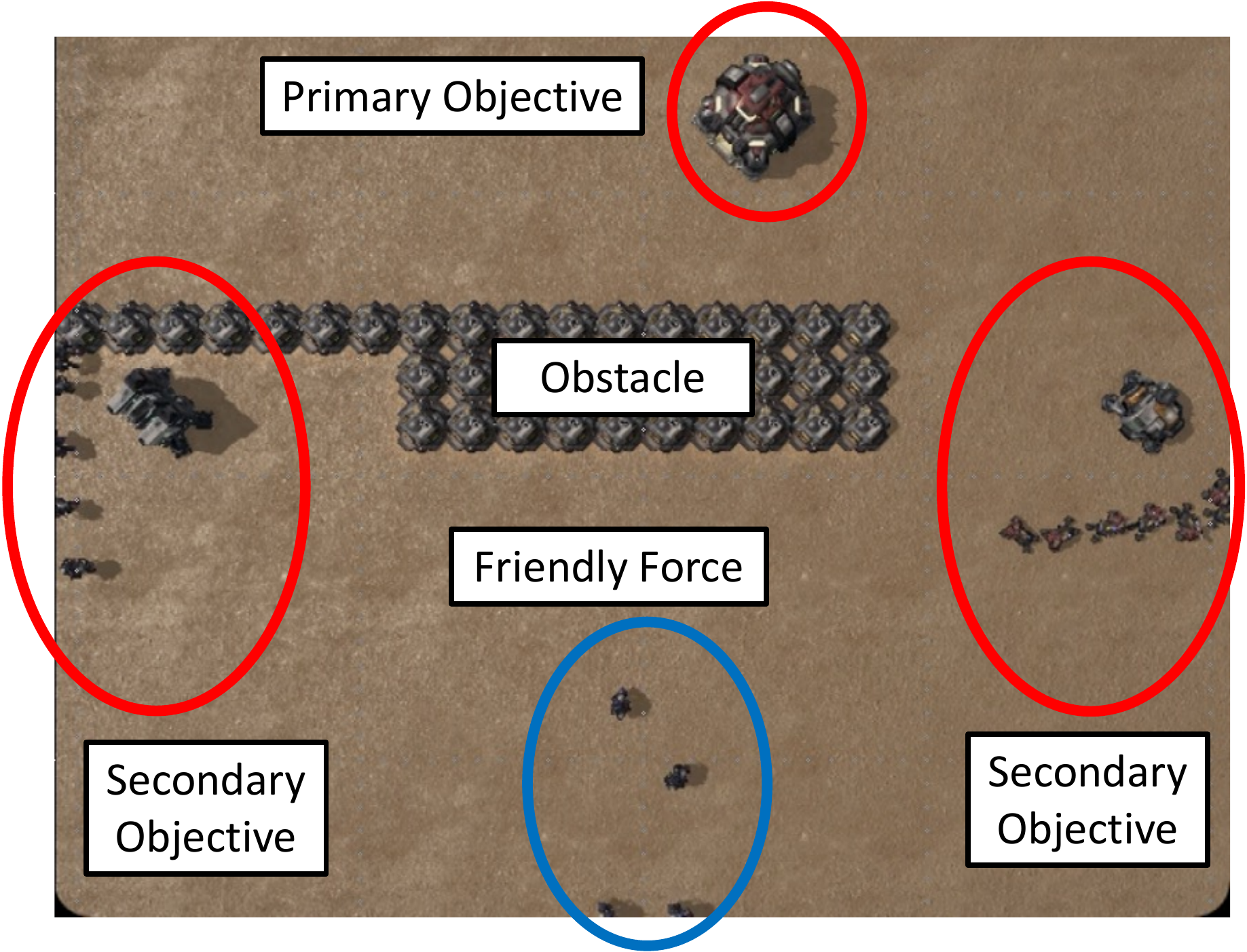}
    \caption{An example scenario generated from our combat task implemented in the StarCraft~II game engine.}%
	\label{Fig:SC2Scenario}
\end{figure}
% ......................................................................................

Our learning problem is a force-on-force combat task implemented in the SC2 game engine. An example scenario is depicted in Fig.~\ref{Fig:SC2Scenario}. The agent controls the Friendly force, and its \emph{primary objective} is to destroy a \emph{CommandCenter} (CC) building belonging to the Enemy force. Both the Friendly and Enemy forces consist of multiple individual \emph{units}, possibly of different \emph{types} (\eg, infantry, tanks, aircraft). The various unit types have different capabilities and differ in their combat effectiveness against other unit types.

Formally, the problem is an episodic MDP $(S, A, P, R, p_0)$, where $S$ is the set of states, $A$ is the set of actions, $P(s'|s, a)$ is the transition probability distribution, $R(s)$ is the reward function, and $p_0(s)$ is the initial state distribution. The state consists of the type, location, allegiance, and ``health'' of each unit on the map. The agent observes the state as a \emph{pixel map}, which is a tensor of size $H\times W\times C$ where each ``pixel'' is a vector of length $C$ giving the properties of the unit at the corresponding location (if any). We refer to a sampled initial state $s_0 \sim p_0$ as a \emph{scenario}.

The process of sampling a scenario proceeds as follows. The Friendly force is created at the bottom of the map by sampling a unit type from $\{\text{\emph{Marine}}, \text{\emph{Marauder}}\}$, then spawning a number of units of that type such that their total \emph{resource cost}%
\footnote{The resource cost is the amount of \emph{Minerals} plus \emph{Vespene Gas} required to construct the unit in the SC2 game. The cost gives a rough measure of the ``strength'' of armies that consist of different types of units.}
is equal to the Friendly \emph{budget}. The Enemy CC spawns at the top of the map at a randomized position. With probability $0.5$, the CC is defended by two \emph{SiegeTankSieged} units, which are like emplaced artillery that can deal great damage to infantry from long range. Additional Enemy reinforcements of type \emph{Hellbat} spawn near the CC with probability $0.5$ at a time step sampled uniformly from $[20, 100]$.%
\footnote{This technically makes the MDP non-stationary, although we consider only stationary policies.}
The map is divided conceptually into three vertical ``lanes'': left, middle, and right. Each lane is blocked by impassible terrain independently with probability $0.5$. Two \emph{secondary objectives} are placed halfway up the map in the left and right lanes, on the Friendly side of any terrain. Each secondary objective consists of a building of a type chosen without replacement from $\{\text{\emph{Barracks}}, \text{\emph{EngineeringBay}}, \text{\emph{Factory}}, \text{\emph{Starport}}\}$, and a defending formation of Enemy units of a type chosen without replacement from $\{\text{\emph{Marauder}}, \text{\emph{Marine}}, \text{\emph{Hellion}}\}$ %in a number corresponding to a random budget, and 
that is placed near the building. % in a randomized spatial arrangement. 
If all defenders of a secondary objective are destroyed, the corresponding building disappears %from the map 
and is replaced by Friendly reinforcements whose type is determined by the type of the building: \emph{Barracks} give either \emph{Marauders} or \emph{Marines} with equal probability, the \emph{EngineeringBay} gives no reinforcements, the \emph{Factory} gives \emph{SiegeTanks}, and the \emph{Starport} gives \emph{Banshees}. %Notably, the Friendly force can obtain \emph{Banshees} and \emph{SiegeTanks} only by capturing the corresponding secondary objective.

The agent's actions are factored over the possible Friendly unit types: a joint action is a tuple $a = (a_1, \ldots, a_4)$ assigning separate actions to all units of each of the $4$ possible types: \emph{Banshee}, \emph{Marauder}, \emph{Marine} and \emph{SiegeTank}. Actions assigned to unit types that do not exist at a certain timestep are ignored. The possible actions for each type of unit are \emph{MoveGrid(loc)}, \emph{AttackMove(loc)}, \emph{Target(type)} and \emph{NoOp}. The \emph{MoveGrid} action causes units to move to the specified location while ignoring enemies. \emph{AttackMove} causes units to move to a location while engaging any enemies they encounter. \emph{Target} causes units to move toward and engage the closest enemy of the specified type while ignoring other types of enemies. The possible \emph{location} arguments are the cells of a coarse $4\times6$ grid overlaid on the map.

The transition function $P(s'|s, a)$ is defined implicitly by the SC2 game engine.%
\footnote{The SC2 engine is ``close to'' deterministic but there is some uncontrollable stochasticity in the behavior of individual units that causes the precise outcomes of battles to be stochastic.}
The reward function $R$ assigns a large reward for destroying the CC, rewards (costs) for each destroyed Enemy (Friendly) unit proportional to its resource cost, and a fixed cost of $-1$ per timestep. An episode ends when the CC is destroyed, the Friendly force is destroyed, or the time limit of $250$ steps is exceeded.

We designed the learning task so that agents can learn distinguishable strategies that are appropriate in different scenarios. Depending on the types and numbers of Friendly and Enemy units, the Friendlies may be at an advantage or disadvantage when attacking certain objectives. The secondary objectives are natural \emph{sub-goals}: the agent may need to capture them to increase its army strength or gain access to units with necessary capabilities before attacking the primary objective. In particular, one type of secondary objective---the \emph{Starport}---grants reinforcements of type \emph{Banshee} when captured. Banshees are aircraft that ignore terrain, and the possible types of Enemy units specifically exclude anti-air capabilities, meaning that Banshees are invulnerable. Thus, when the Starport is present, capturing it guarantees that the Friendly force can achieve its primary objective, which is the only way to do so if all three lanes are blocked. On the other hand, if the center lane is unobstructed, the Friendlies can attack the CC directly, bypassing the secondary objectives. If the CC is defended, however, the agent might first need to capture a secondary objective to gain reinforcements. Finally, the left and right lanes being blocked means that sometimes capturing a secondary objective requires a detour from the route to the primary objective, which changes the cost-benefit calculus.

% ------------------------------------------------------------------------
\subsection{Expert Agent}%
\label{Sec:ExpertDef}

To help validate our approach to strategy discovery we implemented an \emph{expert policy}, for which we know its ``true'' strategy, based on its decision rules. The expert acts according to the following prioritized list of actions: % (\ie, it executes the first applicable rule):
\begin{enumerate}
    \item If the center lane is unobstructed and the CC is undefended, attack the CC directly.
    \item If the Friendly force has Banshee units, attack the CC with only the Banshee units.
    \item If a Starport is present and the Friendly force is stronger%
    \footnote{Relative combat strength is determined via a heuristic scoring function that accounts for the number of units in each force weighted by a multiplier that quantifies how favorable each type-vs-type match-up is, estimated based on domain knowledge.}
    than the Enemy force defending it, attack the Starport.
    \item If there is a clear path (no terrain or secondary objective) from the Friendly force to the CC and the Friendly force is stronger than the Enemy force defending the CC, attack the CC.
    \item If there is a secondary objective whose defending force is weaker than the Blue force, attack that objective.
    \item Otherwise do nothing.
\end{enumerate}

% ------------------------------------------------------------------------
\subsection{Reinforcement Learning Agent}
Our RL agent consists of a neural network policy trained using the VTrace off-policy actor-critic learning rule \citep{espeholt2018impala}. The network architecture is a modification of the FullyConv architecture introduced by \citet{vinyals2017starcraft}.%
\footnote{We implemented the model in TensorFlow v1.13, building upon the open source code of \citet{reaver}. We used the TensorFlow implementation of the Adam optimizer \citep{kingma2014adam} for parameter learning.}
The policy network consists of a stack of convolutional layers followed by two ``heads'' that output: an estimate of the state value $V(s)$, and the action probability distribution. The latter is factored into three parts that output, for each Friendly unit type, the probabilities of: 1) the possible action types (\emph{AttackMove}, \emph{MoveGrid}, \emph{Target} and \emph{NoOp}); 2) the possible \emph{location} arguments for \emph{AttackMove} and \emph{MoveGrid}; and 3) the possible \emph{type} arguments for \emph{Target}.%The probability of an action $\text{\emph{Func}}(\text{\emph{Arg}})$ is the probability of the action function \emph{Func} times the probability of the applicable argument \emph{Arg}. Actions for unit types that do not exist are ignored when calculating the probability of the joint action.

We trained multiple RL agents, varying the action space, the learning rate, and the ``entropy bonus'' coefficient in the VTrace rule which encourages exploration. While the best RL policies achieve the goal regularly and exhibit non-trivial behavior, they failed to reach the level of performance of the expert policy. Exploration seems to be challenging in this problem, with most training runs requiring millions of training steps before beginning to reach the primary objective consistently. Since our goal is to characterize the strategies in agent policies, regardless of their performance, we used the best of the 
%RL policy 
RL policies to serve as our RL agent. 
%It is possible that different policy architecture choices would produce better performance. It is important to note that our strategy discovery methods are agnostic to the quality of the policy being analyzed, and in fact low-performance policies are in some ways more challenging to analyze than high-performance policies, since the poor performance is often the result of less-coherent behavior. 
As explained in Sec.~\ref{Sec:logic}, we need traces from a random agent during strategy inference for comparing against the policy (RL or expert) agent's behavior. To that end, we also designed an agent that, at each step, selects actions uniformly at random.

% ------------------------------------------------------------------------
\subsection{High-Level Feature Extractor}

% why do we need high-level features
The SC2 engine provides structured observations that provide agents with approximately the same information that a human player would see given a static snapshot of the game's screen---\eg information about which units are present on the screen, their game properties, or their location on the game board. To extract interpretable logical formulas, however, we need a higher-level, semantic representation that allows us to describe the state of the environment and the Friendly and Enemy forces' behavior, including their dynamics, in a strategically meaningful fashion. 

% ......................................................................................
\begin{table*}[!tb]
	\centering
	\caption{The unit groups defined for the high-level feature extractor that were used in our experiments.}%
	\label{Table:UnitGroups}
	\begin{tabular}{l L{185pt} L{150pt}}
	\toprule
	\textbf{Unit Group} & \textbf{Description} \\ 
	\hline
	\texttt{Blue}   & all unit types belonging to the Friendly force & \texttt{Marauder, Marine, SiegeTank, Banshee}\\
	\hdashline[.5pt/1pt]
	\texttt{Red}    & all unit types belonging to the Enemy force & \texttt{CommandCenter, Hellbat, Hellion, Marauder, Marine, SiegeTankSieged}\\
	\hdashline[.5pt/1pt]
	\texttt{Air}    & aerial unit types, \ie that move through the air & \texttt{Banshee} \\
	\hdashline[.5pt/1pt]
	\texttt{Ground} & ground unit types & \texttt{CommandCenter, Marine, SiegeTank, SiegeTankSieged, Hellbat, Marauder, Hellion}\\
	\hdashline[.5pt/1pt]
	\texttt{Mobile} & unit types that can move through the game board, \ie are not static & \texttt{Marine, SiegeTank, Hellbat, Marauder, Hellion}\\
	\hdashline[.5pt/1pt]
	\texttt{ProductionFacility} & unit types corresponding to buildings that generate new units & \texttt{Barracks, EngineeringBay, Factory, Starport}\\
	\hdashline[.5pt/1pt]
	\texttt{Obstacle} & unit types that serve as immobile barriers & \texttt{SupplyDepot}\\
	\bottomrule
	\end{tabular}
\end{table*}
% ......................................................................................

% feature extractor info, unit groups concept
One option would be to train classifiers to identify features of interest from labeled datasets. However, given the richness and simplicity of the information provided by the SC2 engine, we instead implemented a high-level SC2 feature extractor%
\footnote{Our SC2 feature extractor is available at: \url{https://github.com/SRI-AIC/sc2-feature-extractor}.}
whose input is the data provided by the SC2 ``raw'' features and output is a set of categorical features (as depicted in Fig.~\ref{Fig:Framework}). Some features describe only the current state of the world while others capture dynamics occurring between consecutive timesteps. The feature extractors are built around the notion of \emph{unit groups}, each defining a set of unit types for which a certain feature is computed. %, \ie it functions as a filter for the units within some force. 
Table~\ref{Table:UnitGroups} lists the unit groups used in our experiments. In addition, we created a group for each unit type \ie containing a single element, with the goal of assessing the capability of Strategy Inference generalizing among groups of units.

% feature extractors list
We implemented the following high-level feature extractors, each generating a set of features at each discrete timestep:
\begin{description}
    \item[Unit group presence:] detects the presence (or absence) of Friendly and Enemy unit groups. It outputs Boolean features of the form \texttt{\{FORCE\}\_\{GROUP\}}, where \texttt{\{FORCE\}} takes the value of either \texttt{Friendly} or \texttt{Enemy}, and \texttt{\{GROUP\}} is the unit group of the type we want to detect. These features are $\True$ if \emph{at least one} unit of a type in the group belonging to the corresponding force is present (alive) in the game, and $\False$ otherwise.
    \item[Defender presence:] detects the presence of defender units around some task objective and outputs features in the form \texttt{Defender\_\{DEFENDED\_G\}\_\{DEFENDER\_G\}}, where \texttt{\{DEFENDED\_G\}} is the unit (building) group being defended, and \texttt{\{DEFENDER\_G\}} is the group defending the building. The output Boolean features are $\True$ if at least one unit of each group (defender, defended) is present in the game and $\False$ otherwise.
    \item[Distance to Enemy:] computes the relative distance between given Friendly and Enemy unit groups, where the distance between groups corresponds to the minimal distance between \emph{any pair of units} belonging to each group. It outputs categorical features%
    \footnote{For purposes of Trace Clustering and Strategy Inference, categorical features are converted to Boolean via one-hot encoding, where the new features' names include the original features' value/label, \eg \texttt{Distance\_Blue\_Red} originates \texttt{Distance\_Blue\_Red=Melee/Close/Far/Undefined}. }
    in the form \texttt{Distance\_\{FRIENDLY\_G\}\_\{ENEMY\_G\}}, where the value is selected by comparing the ratio of that distance to the maximal game board distance against predefined threshold values.%
    \footnote{The feature extractor's parameters presented in this paper were selected empirically based on whether the resulting features provided an accurate description of the game aspect they are trying to capture.}
    Specifically, denoting the distance ratio by $d$, the features have a value of \texttt{Undefined} if no units are present belonging to either unit group, \texttt{Melee} if $d\leq0.05$, \texttt{Close} if $d\leq0.1$, and \texttt{Far} otherwise. 
    \item[Relative cost:] compares the cost of a Friendly group of units relative to that of an Enemy group, where group cost is measured as the sum of the SC2 game costs of all units whose types belong to that group. Denoting by $r$ the ratio between the Friendly and Enemy groups' cost, this extractor outputs categorical features in the form \texttt{RelativeCost\_\{FRIENDLY\_G\}\_\{ENEMY\_G\}}, where the features are \texttt{Undefined} if no units are present belonging to either group, \texttt{Disadvantage} if $r<0.9$, \texttt{Advantage} if $r^{-1}<0.9$, and \texttt{Balanced} otherwise.
    \item[Under attack:] detects whether a Friendly or Enemy group is under attack by monitoring the group units' total health. It outputs Boolean features in the form \texttt{UnderAttack\_\{FORCE\}\_\{GROUP\}}, where the feature is $\True$ if the sum of the group units' health has decreased in relation to the previous timestep, and $\False$ otherwise.
    \item[Relative movement:] detects the movement of Friendly and Enemy forces relative to each other. The movement is calculated by computing the velocity, measured between the current and previous timestep, of the forces' center-of-mass, where this location is only considered for groups that are present at \emph{both} timesteps. This extractor computes two types of features: \texttt{Advancing\_\{FORCE\}\_\{FORCE\_G\}\_\{OTHER\_G\}}, detecting whether a group belonging to either the Friendly or the Enemy force is moving towards the other; and \texttt{Retreating\_\{FORCE\}\_\{FORCE\_G\}\_\{OTHER\_G\}}, detecting whether the groups are moving away from each other. The movement of forces relative to each other, \ie, advance and retreat, is controlled by the angle between a forces' movement direction and the vector defined by the forces' center-of-mass relative to each other. Denoting the value of this angle in radians by $\alpha$, the \texttt{Advancing} features have a Boolean value according to whether $\alpha\in[0,\alpha_\textrm{mov}[$, while the \texttt{Retreating} features have a Boolean value according to whether $\alpha\in]\pi-\alpha_\textrm{mov},\pi]$, with $\alpha_\textrm{mov}=1.15$ in our experiments. 
    \item[Between:] detects whether an Enemy ``barrier'', formed by a group of units, is between Friendly and Enemy groups. It outputs Boolean features of the form \texttt{Between\_\{BARRIER\_G\}\_\{FRIENDLY\_G\}\_\{ENEMY\_G\}}. Let $\boldsymbol{\alpha}$ denote the angles between the vectors formed by the location of one unit belonging to the \texttt{\{FRIENDLY\_G\}} group and the location of a unit of the \texttt{\{ENEMY\_G\}} group, relative to all units of the \texttt{\{BARRIER\_G\}} group. These features are $\True$ if $\sum_{\alpha\in\boldsymbol{\alpha}}{\boldsymbol{1}\left[\alpha<\alpha_{btw}\right]}>0.25\card{\alpha}$ and $\False$ otherwise, where $\boldsymbol{1}$ is the indicator function and $\alpha_{btw}=0.1$, \ie a barrier is considered to exist between the Friendly and Enemy groups if at least $25\%$ of the pairs of units have a barrier unit type intersecting the line connecting their locations.
    \item[Agent's actions:] as described earlier, our agents use high-level actions to command the units in the SC2 scenario. This extractor outputs Boolean features for each type of agent action, \ie \texttt{\{ACTION\}\_Friendly\_\{G\}} that are $\True$ whenever a unit of a type belonging to group \texttt{\{G\}} has performed the corresponding \texttt{\{ACTION\}}, and $\False$ otherwise, with $\texttt{\{ACTION\}}\in\{\texttt{MoveGrid},\texttt{AttackMove},\texttt{Target},\texttt{NoOp}\}$. For the target actions, the extractor generates features in the form \texttt{Target\_\{FRIENDLY\_G\}\_\{ENEMY\_G\}}, that are $\True$ whenever a unit of the \texttt{\{FRIENDLY\_G\}} group targets a unit whose type belongs to \texttt{\{ENEMY\_G\}}.
\end{description}

% ------------------------------------------------------------------------
\subsection{Experimental Setup}

% ......................................................................................
\begin{table*}[!tb]
	\centering
	\caption{The set of high-level features considered in our experiments.}%
	\label{Table:Features}
	\begin{tabular}{l l }
	\toprule
	\textbf{Condition Features} & \textbf{Action-Related Features}\\
	\hline
	\texttt{Between\_Obstacle\_Blue\_CommandCenter} & \texttt{Advancing\_Friendly\_Air\_CommandCenter}\\
	\texttt{Between\_Red\_Blue\_CommandCenter} & \texttt{Advancing\_Friendly\_Blue\_CommandCenter}\\
	\texttt{Between\_Red\_Blue\_ProductionFacility} & \texttt{Advancing\_Friendly\_Blue\_ProductionFacility}\\
	\texttt{Defender\_CommandCenter\_Red} & \texttt{Advancing\_Friendly\_Blue\_Red}\\
	\texttt{Defender\_EngineeringBay\_Red} & \texttt{Advancing\_Friendly\_Blue\_Starport}\\
	\texttt{Defender\_Factory\_Red} & \texttt{Advancing\_Friendly\_Ground\_CommandCenter}\\
	\texttt{Defender\_ProductionFacility\_Red} & \texttt{Advancing\_Friendly\_Ground\_Mobile}\\
	\texttt{Defender\_Starport\_Red} & \texttt{Advancing\_Friendly\_Ground\_ProductionFacility}\\
	\texttt{Distance\_Blue\_CommandCenter} & \texttt{Advancing\_Friendly\_Ground\_Starport}\\
	\texttt{Distance\_Blue\_ProductionFacility} & \texttt{AttackMove\_Friendly\_Air}\\
	\texttt{Present\_Enemy\_CommandCenter} & \texttt{AttackMove\_Friendly\_Blue}\\
	\texttt{Present\_Enemy\_EngineeringBay} & \texttt{AttackMove\_Friendly\_Ground}\\
	\texttt{Present\_Enemy\_Factory} & \texttt{MoveGrid\_Friendly\_Air}\\
	\texttt{Present\_Enemy\_Hellbat} & \texttt{MoveGrid\_Friendly\_Blue}\\
	\texttt{Present\_Enemy\_Hellion} & \texttt{MoveGrid\_Friendly\_Ground}\\
	\texttt{Present\_Enemy\_Marauder} & \texttt{NoOp\_Friendly\_Air}\\
	\texttt{Present\_Enemy\_Marine} & \texttt{NoOp\_Friendly\_Blue}\\
	\texttt{Present\_Enemy\_Obstacle} & \texttt{NoOp\_Friendly\_Ground}\\
	\texttt{Present\_Enemy\_ProductionFacility} & \texttt{Retreating\_Friendly\_Blue\_Red}\\
	\texttt{Present\_Enemy\_SiegeTankSieged} & \texttt{Target\_Air\_CommandCenter}\\
	\texttt{Present\_Enemy\_Starport} & \texttt{Target\_Ground\_CommandCenter}\\
	\texttt{Present\_Friendly\_Air} & \texttt{Target\_Ground\_Hellbat}\\
	\texttt{Present\_Friendly\_Blue} & \texttt{Target\_Ground\_Hellion}\\
	\texttt{RelativeCost\_Blue\_Red} & \texttt{Target\_Ground\_Marauder}\\
	\texttt{UnderAttack\_Friendly\_Ground} & \texttt{Target\_Ground\_Mobile}\\
	 & \texttt{Target\_Ground\_SiegeTankSieged}\\
	\bottomrule
	\end{tabular}
\end{table*}
% ......................................................................................

% scenario sampling, trace generation
We followed the pipeline outlined in Fig.~\ref{Fig:Framework} and started by generating $1{,}000$ replays in our SC2 task for each of the three agents, \ie \emph{expert}, \emph{RL} and \emph{random}. We sampled the scenario parameters uniformly at random for each episode, resulting in $1{,}000$ unique starting conditions, \ie scenarios, from which the game then evolved in a different manner for each agent according to its policy, resulting in a total of $3{,}000$ distinct agent trajectories. 

% feature extraction
We then used the high-level feature extractor to generate the set of agent traces. We used a total of $51$ high-level features, as listed in Table~\ref{Table:Features}. We parameterized the extractors' unit groups based on domain knowledge and the expert policy to produce the minimum number of features necessary for a meaningful description of the agents' strategies. As seen in Table~\ref{Table:Features}, for the conditions we focused on features denoting the presence of all possible types of units, including specific secondary objectives, whether the objectives are defended and whether there are obstacles or Enemy units between the Friendly force and the relevant targets. For the action-related features, we included the detection of the agent's movement toward important Enemy units, and detectors for the various actions of the agent for different unit groups.

% feature conversion
% As described in Sec.~\ref{Sec:Framework}, our framework assumes the existence of a set of Boolean features from which to extract the agent traces prior to Trace Clustering and Strategy Inference. Since some of our high-level features are categorical, we converted them to Boolean features via one-hot encoding, where the new features' names include the original features' value/label, \eg \texttt{Distance\_Blue\_CommandCenter} originates \texttt{Distance\_Blue\_CommandCenter=Melee/Close/Far/Undefined}. 

% trace embedding and clustering
After extracting the agent traces, we randomly selected $90\%$ of the data for Trace Clustering and Strategy Inference, while the remaining $10\%$ of traces were used for evaluation purposes. We then generated the embedding representation for each trace as detailed in Sec.~\ref{Subsec:Clustering}, where we used the condition features listed in Table~\ref{Table:Features} to get feature set $\F_{task}$ from which we derived the embedding $\E_{fc}$, and the action-related features to get feature set $\F_{act}$ from which we derived the embedding $\E_{sgt}$. For the computation of the SGT embedding we used $\kappa=1$ as suggested in \citep{ranjan2016sequence} to capture long-term dependencies between features, and for the feature counts embedding we used a discount factor of $\gamma=0.99$. The embedding representation for each trace was generated by concatenating the two embeddings as specified by Eq.~\ref{Eq:Embed}. This resulted in an embedding size of $\card{\E}=\card{\E_{fc}}+\card{\E_{sgt}}=60+4{,}119=4{,}179$ with an alphabet size for the SGT embedding of $\card{\V}=72$ for the expert agent traces, and $\card{\E}=60+5{,}305=5{,}365$ and $\card{\V}=78$ for the RL agent traces, where we filtered the features that had constant values in all timesteps of all traces. We then clustered the traces using HAC using the cosine distance and the \emph{complete} linkage criterion, which minimizes the maximum distance between the embeddings of pairs of traces belonging to different clusters.

% strategy discovery
For strategy discovery, we used the strategy templates described in Sec.~\ref{Sec:StrategyDiscovery}, with the following ranges for the parameters: action duration $d\in\set{0, 2, 5, 10, 20, 30, 40, 50, 60, 70, 80, 90, 100, 150, 200}$; satisfaction rate $r\in\set{0.7,0.8,0.9,1.0}$. For conditions $C$ and goals $G$ we used the atomic propositions from the condition features in Table~\ref{Table:Features}, as well as their negations, whereas for actions $A$ we used the action-related features. %
%
% \footnote{In general, the conditions and goals could be more complex Boolean formulas over the feature space.}
%
Having identified the range of each parameter, we generate all possible strategies according to the specified templates.

% RESULTS
% =========================================
\section{Results and Analysis}%
\label{Sec:Results}

We now detail the results for each component in our framework for the SC2 experiments just described.

% ------------------------------------------------------------------------
\subsection{Agent Comparison}

% ......................................................................................
\begin{table*}[!tb]
	\centering
	\caption{Comparison of results for the traces produced for each agent. ``\% Episodes'' is the percentage of episodes that ended with the corresponding event being true, while ``\% Time'' refers to the percentage of all the timesteps (across all episodes) in which the event is true.}%
	\label{Table:AgentComparison}
	\begin{tabular}{l R{65pt} R{65pt} R{65pt} }
	\toprule
	 & \textbf{Expert} & \textbf{RL} & \textbf{Random}\\
	\hline
	{Num. episodes/traces} & {1,000} & {1,000} & {1,000}\\
	{Mean trace length} & {$152.44\pm73.21$} & {$214.41\pm63.31$} & {$162.77\pm80.52$}\\
	{\% Episodes that timed-out} & {$29$} & {$64$} & {$34$}\\
	{\% Episodes CC destroyed} & {$52$} & {$13$} & {$4$}\\
	{\% Ep. Starport destroyed (when present)} & {$31$} & {$25$} & {$6$}\\
	{\% Ep. both sec. objectives destroyed} & {$4$} & {$5$} & {$1$}\\
	{\% Ep. with no agent units left} & {$19$} & {$24$} & {$63$}\\
	{\% Time advancing towards Mobile} & {$5$} & {$31$} & {$13$}\\
	{\% Time advancing towards CC} & {$8$} & {$28$} & {$14$}\\
	{\% Time Friendly under attack} & {$10$} & {$10$} & {$12$}\\
	{\% Time Enemy under attack} & {$37$} & {$18$} & {$15$}\\
	{\% Time CC under attack} & {$23$} & {$5$} & {$4$}\\
	{\% Time NoOp} & {$60$} & {$63$} & {$68$}\\
	{\% Time Target Enemy} & {$36$} & {$64$} & {$69$}\\
	{\% Time AttackMove} & {$40$} & {$79$} & {$67$}\\
	{\% Time MoveGrid} & {$0$} & {$68$} & {$69$}\\
	\bottomrule
	\end{tabular}
	\vspace{-14pt}
\end{table*}
% ......................................................................................

% comparative analysis between agents, especially expert and RL
Table~\ref{Table:AgentComparison} compares the performance of the three agents across different aspects of the task. As we can see, the expert performs much better than the other agents. It destroys the CC more often than the RL agent (four times as much), as well as the \emph{Starport}---the most important secondary objective. It also seems to apply a more efficient strategy, spending more time attacking the Enemy compared to other agents (see ``\% Time Enemy/CC under attack''), which means it is more selective in choosing which units to attack with. %In addition, the expert traces are shorter on average than those of the RL agent. 
In contrast, the RL agent seems to specialize in attacking the secondary objectives without following through with an attack to the CC. It nevertheless learned a good enough strategy to not be defeated very often---only $24\%$ of episodes end up with no units compared to $63\%$ for the baseline random agent. As discussed earlier, although our RL agent failed to reach the expert's level of performance, our goal is to analyze and understand the behavior of agents, regardless of their performance. Low-performance policies are arguably more challenging to analyze than high-performance policies, since poor performance is often the result of less-coherent behavior. 

% ------------------------------------------------------------------------
\subsection{Trace Clustering Evaluation}

% ------------------------------------------------------------------------
% \subsubsection{Internal Evaluation}

% ......................................................................................
\begin{figure}[!tb]
	\centering
    \begin{subfigure}[b]{0.4\columnwidth}
        \includegraphics[width=\textwidth]{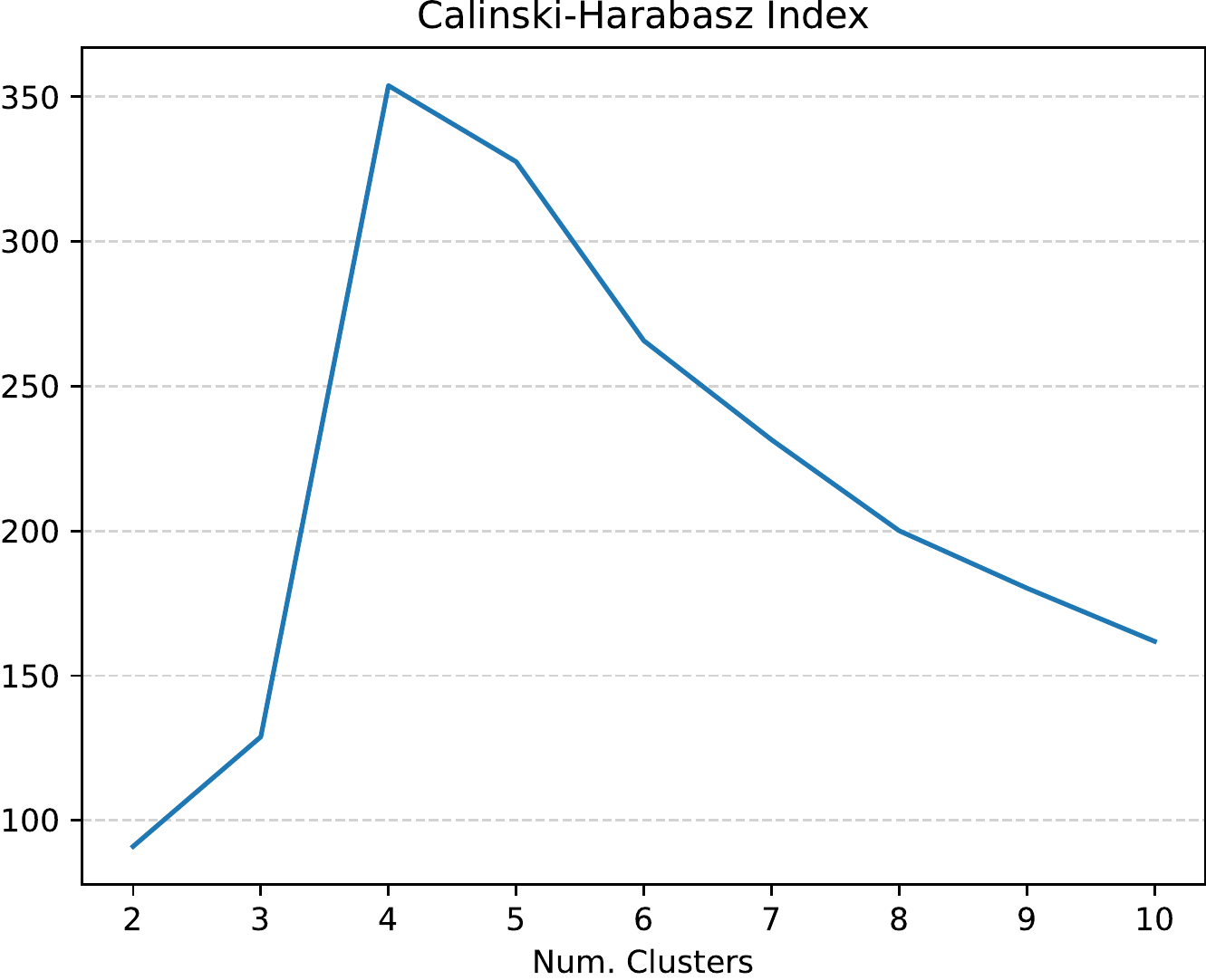}
        \caption{Expert internal score}%
        \label{Fig:ExpertScore}
    \end{subfigure}%
    \hspace{10pt}
    \begin{subfigure}[b]{0.4\columnwidth}
        \includegraphics[width=\textwidth]{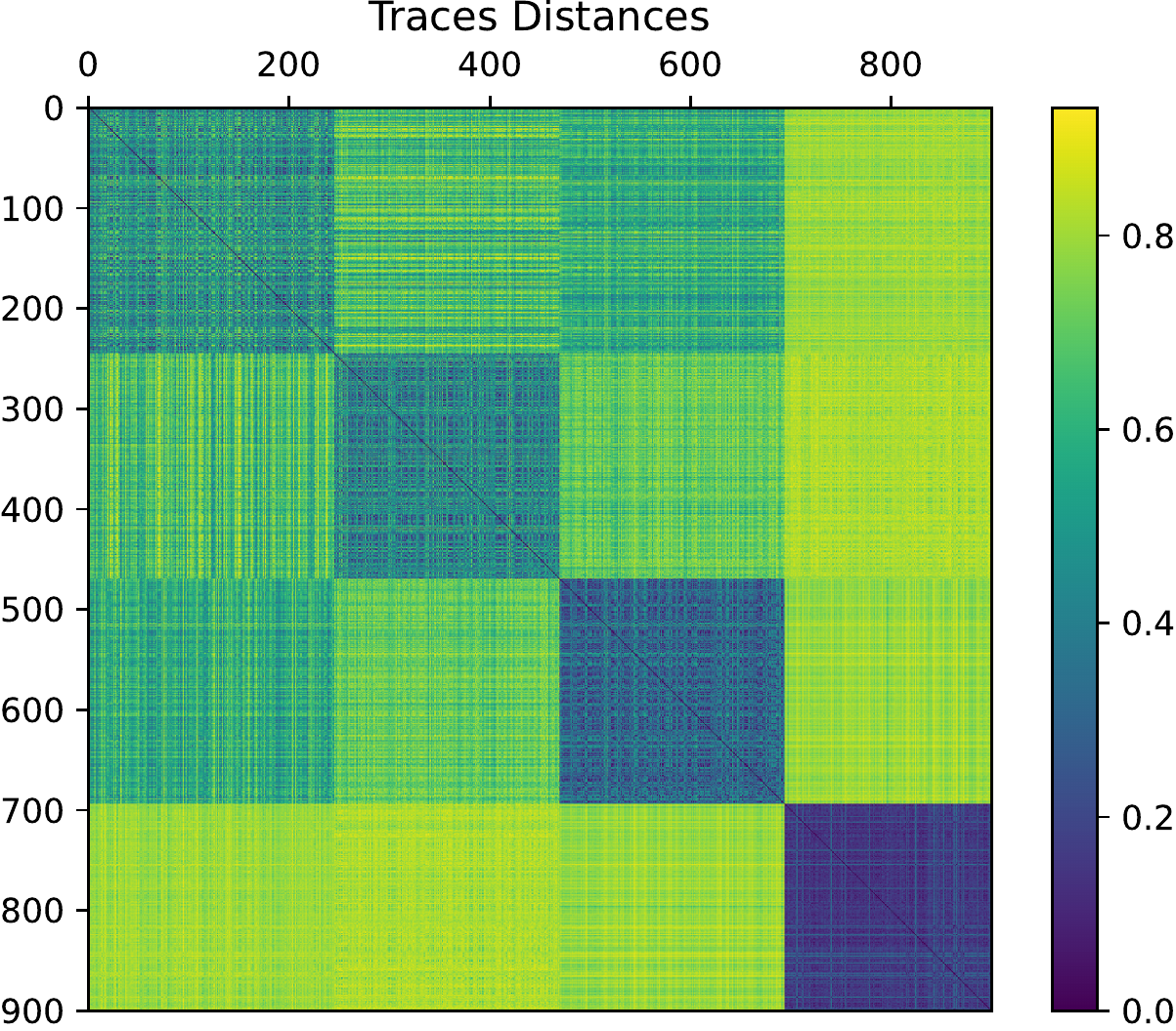}
        \caption{Expert trace distances}%
        \label{Fig:ExpertDist}
    \end{subfigure}
    \begin{subfigure}[b]{0.4\columnwidth}
        \includegraphics[width=\textwidth]{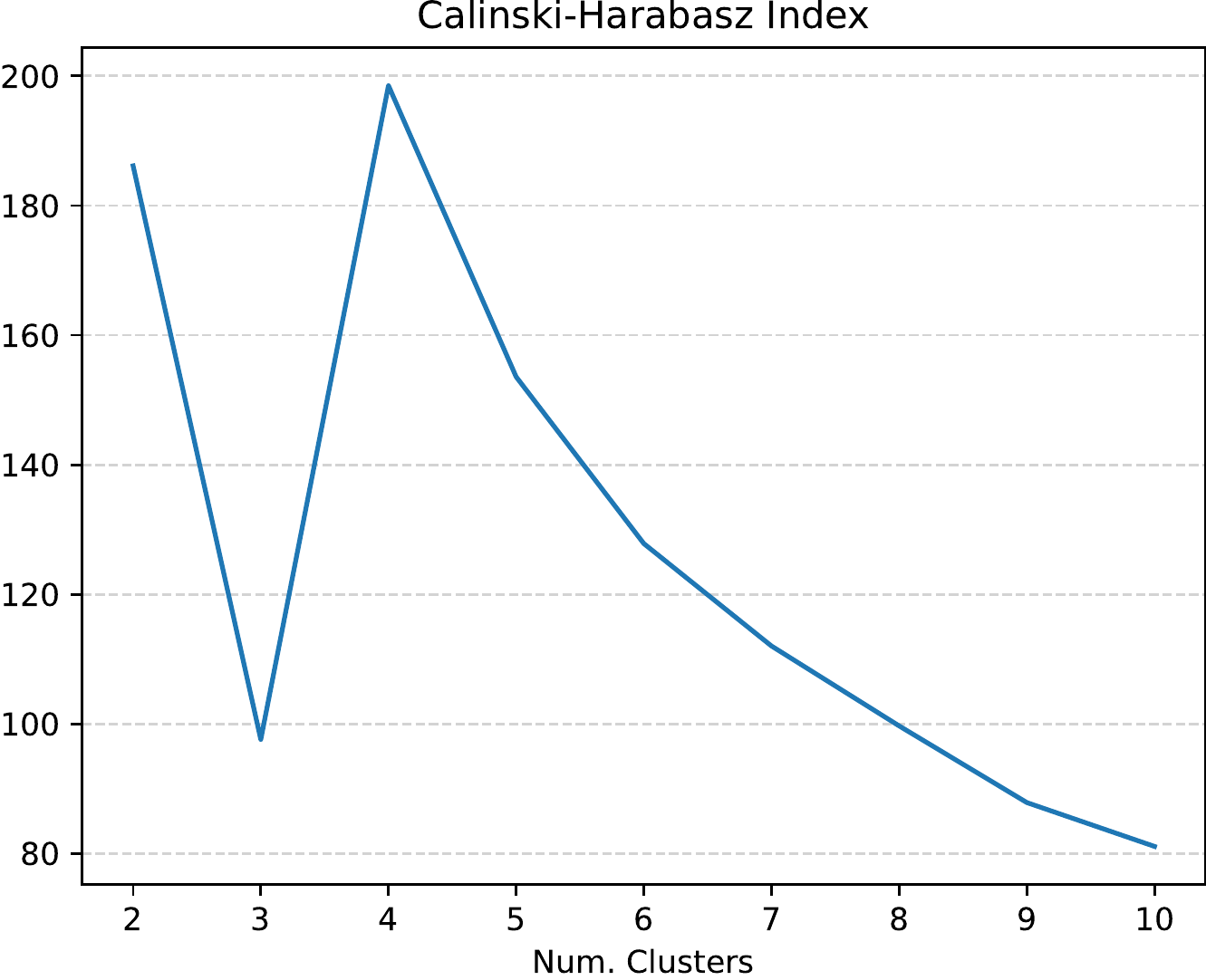}
        \caption{RL internal score}%
        \label{Fig:RLScore}
    \end{subfigure}%
    \hspace{10pt}
    \begin{subfigure}[b]{0.4\columnwidth}
        \includegraphics[width=\textwidth]{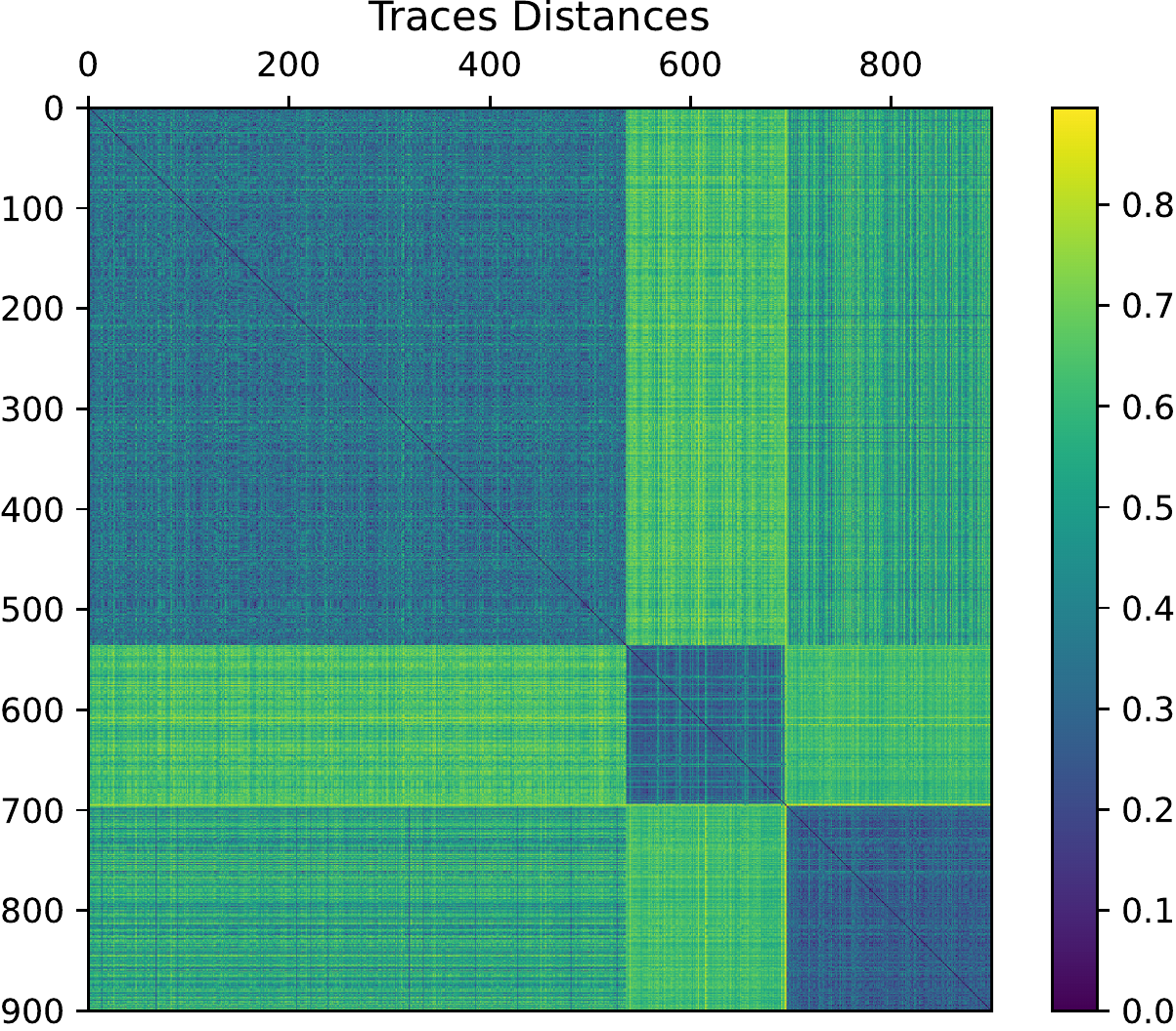}
        \caption{RL trace distances}%
        \label{Fig:RLDist}
    \end{subfigure}
	\caption{Trace clustering results for the expert agent's traces (top) and RL agent's traces (bottom). Left: internal evaluation for different number of clusters (higher is better), right: confusion matrix, for the optimal number of clusters, showing the pairwise distances between the embeddings of all clustered traces ($900$), organized by cluster (darker squares represent more cohesive clusters, lighter colored regions outside of squares represent better separated clusters).}%
	\label{Fig:ClusterResults}
\end{figure}
% ......................................................................................

% clustering results for both agents
Fig.~\ref{Fig:ClusterResults} shows the results of trace clustering. As mentioned earlier, we use HAC for clustering traces, which does not prescribe an optimal number of clusters. We evaluated the different trace partitions from $2$ to $10$ clusters using the Cali\'{n}ski-Harabasz score, which indicated an optimal number of $4$ clusters for both the expert and RL agent (see Figs.~\ref{Fig:ExpertScore} and \ref{Fig:RLScore}). Figs.~\ref{Fig:ExpertDist} and \ref{Fig:RLDist} show the confusion matrices with the pairwise distances between the embeddings of the expert's and RL agent's traces, respectively. 
%Regarding the distribution of traces for each cluster, 
For the expert agent, cluster $0$ has $245$ traces, clusters $1$ and $2$ have $224$ traces each, and cluster $3$ has $207$ traces. For the RL agent, cluster $0$ has $535$ traces, cluster $1$ has $158$ traces, cluster $2$ has $2$ traces, and cluster $3$ has $205$ traces.

% overall analysis
Overall, although the optimal number of clusters is the same for both agents, we see that the traces of the expert agent are more cohesive and better separated. This indicates more consistent behavior, which is expected given the inherently stochastic nature of the RL agent's actions. The distribution of traces over clusters is also more balanced for the expert, meaning that it likely employed more distinct behaviors---we again note that the scenario parameters were sampled uniformly at random for each episode. Overall, this shows that the number of clusters suggested by the internal evaluation criterion can be a good indication of the number of qualitatively different strategies employed by the agent. 

% ------------------------------------------------------------------------
% \subsubsection{Cluster Visualizations}
\subsection{Cluster Visualizations}

% visualizations rationale, detailing
To help understand how the Friendly (\ie the agent's) and Enemy units behave in different clusters for different agents, we designed a visualization tool%
\footnote{Our spatio-temporal visualization tool is provided alongside the SC2 high-level feature extractor.}
%
%that captures 
to capture the spatio-temporal patterns of the units on the game board given a set of agent traces. The tool produces images and animations showing the occupancy of the different units over time, up to a given timestep ratio $t\in[0,1]$. The following visual properties are used to produce the visualizations:
\begin{description}
    \item[Location:] we create plots representing the location of units in the game board using a top-down view. The entire game board is visualized, and each $x,y$ pixel corresponds to the cell with the same coordinates on the board.
    \item[Color map:] we use color maps to distinguish between the Friendly and Enemy units. In particular, a green-to-blue palette is used to represent the location of the Friendly units over time, and a yellow-to-red palette to represent the location of the Enemy units.
    \item[Color value:] within each color map, the color of each $x,y$ pixel on the plot represents the average relative time that units have occupied the corresponding cell on the game board. Because traces have different length, we first normalize each timestep of a trace to the $[0,1]$ interval. %such that values near $0$ correspond to the presence of units in some location occurring on average at the beginning of traces, while values near $1$ correspond to the presence of units near the end of traces. 
    Then, for each location and force, we compute the average normalized time that the corresponding cell was visited by \emph{any} unit of that force across all traces, up to threshold $t$. Lighter colors within each map (green/yellow) represent a lower average time value, \ie earlier in the traces, while darker colors (blue/red) represent higher time values, \ie closer to the end of traces. 
    \item[Transparency:] to avoid visual cluttering, we set the transparency of each pixel in the plot to be proportional to the relative frequency of the corresponding cell's occupancy during the traces. %As such, locations that were infrequently visited by either the Friendly or Enemy units will have a lower value assigned to the alpha channel of the corresponding pixel. 
    We also use transparency to visualize locations that were occupied by both the Friendly and Enemy units such that both colors can be blended and visualized together.
    \item[Time:] we can linearly interpolate the parameter $t$ in the $[0,1]$ interval to get a sequence of images corresponding to the evolution of units' locations in the traces over time. We note that each image frame in the animation still visualizes what occurred up to $t$ in all traces starting from the initial timestep.
\end{description}

% ......................................................................................
\begin{figure}[!tb]
	\centering
    \begin{subfigure}[b]{0.4\columnwidth}
        \includegraphics[width=\textwidth]{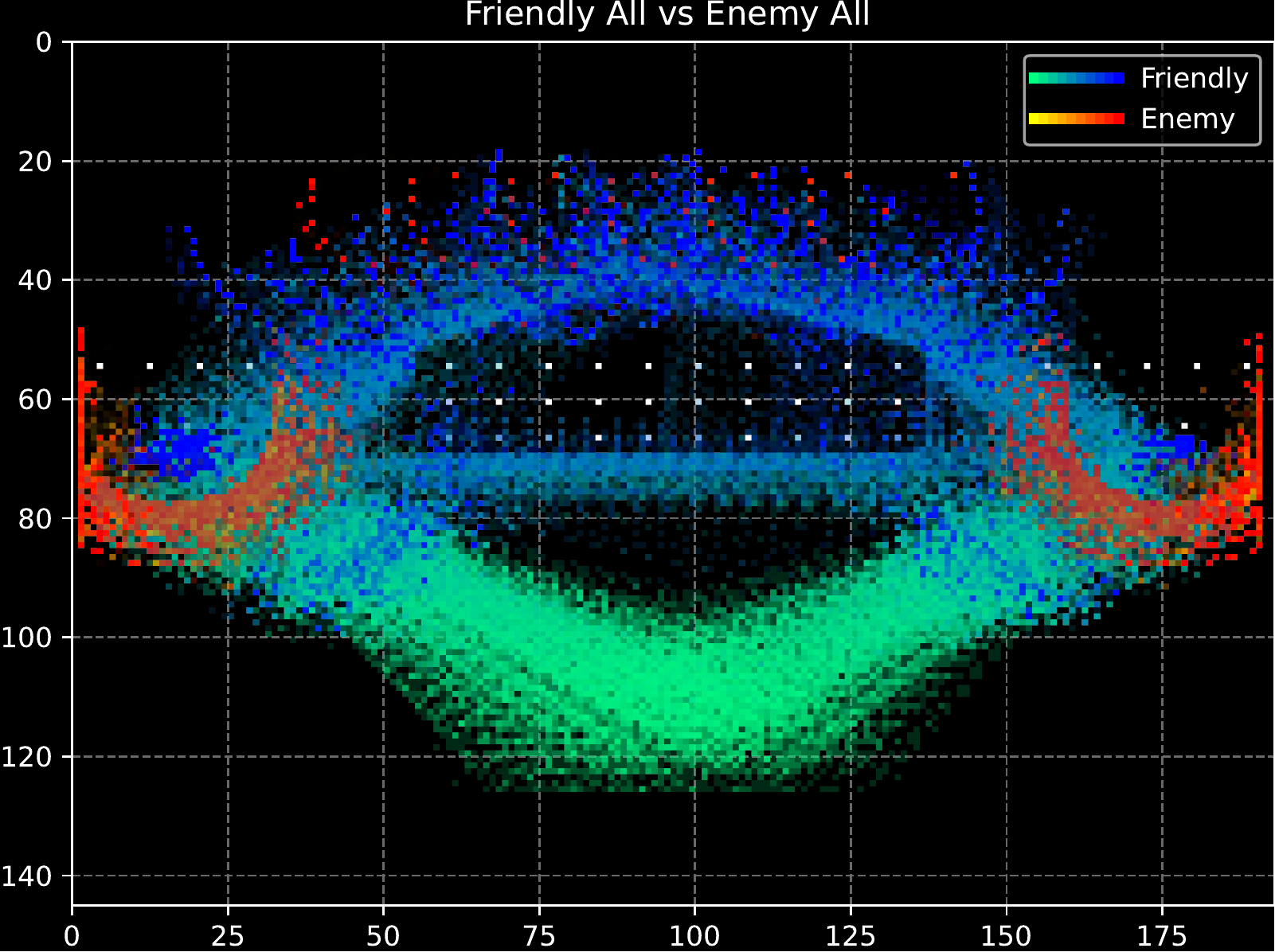}
        \caption{Cluster 0}%
        \label{Fig:Expert0}
    \end{subfigure}%
    \hspace{10pt}
    \begin{subfigure}[b]{0.4\columnwidth}
        \includegraphics[width=\textwidth]{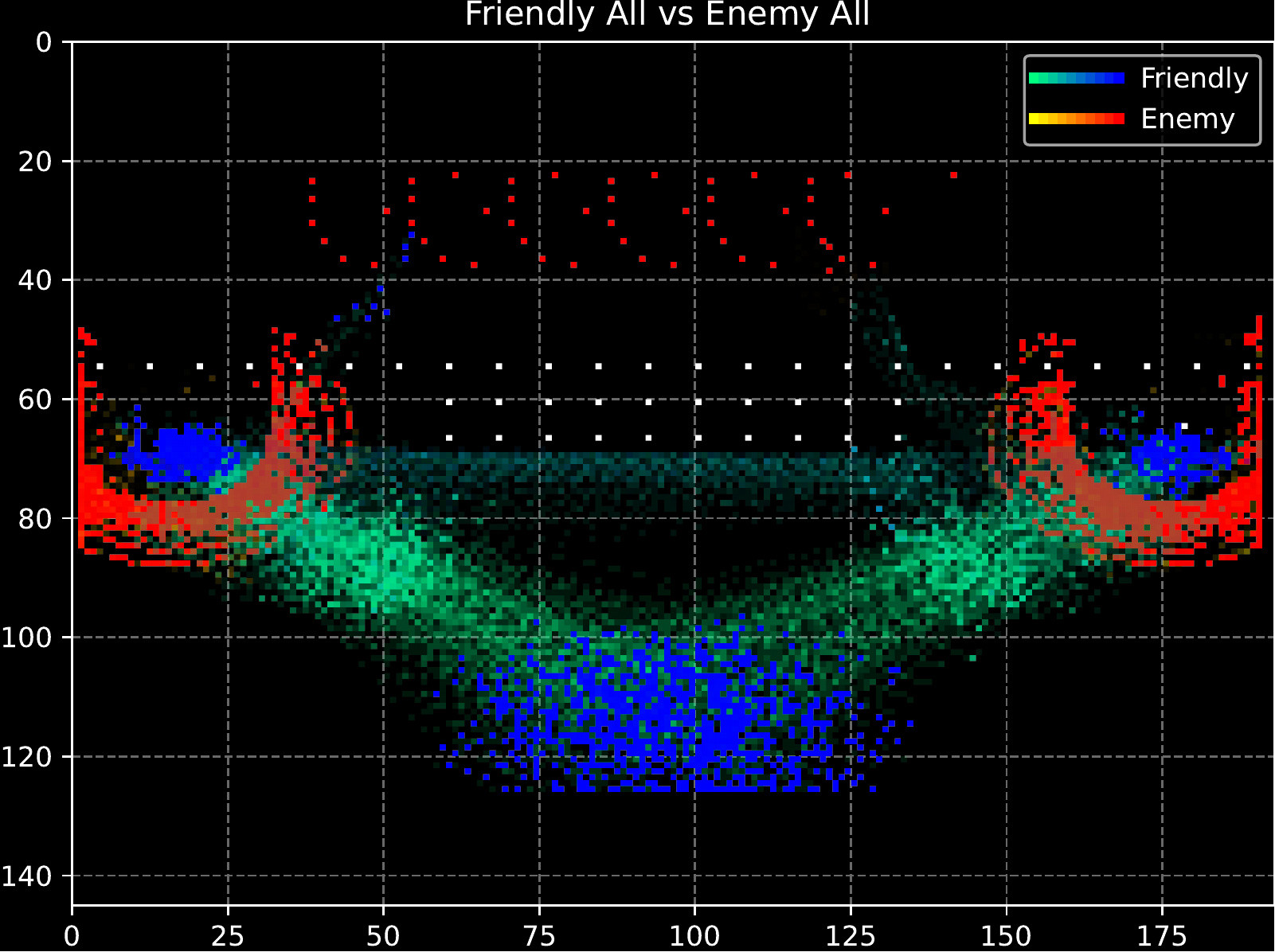}
        \caption{Cluster 1}%
        \label{Fig:Expert1}
    \end{subfigure}
    \begin{subfigure}[b]{0.4\columnwidth}
        \includegraphics[width=\textwidth]{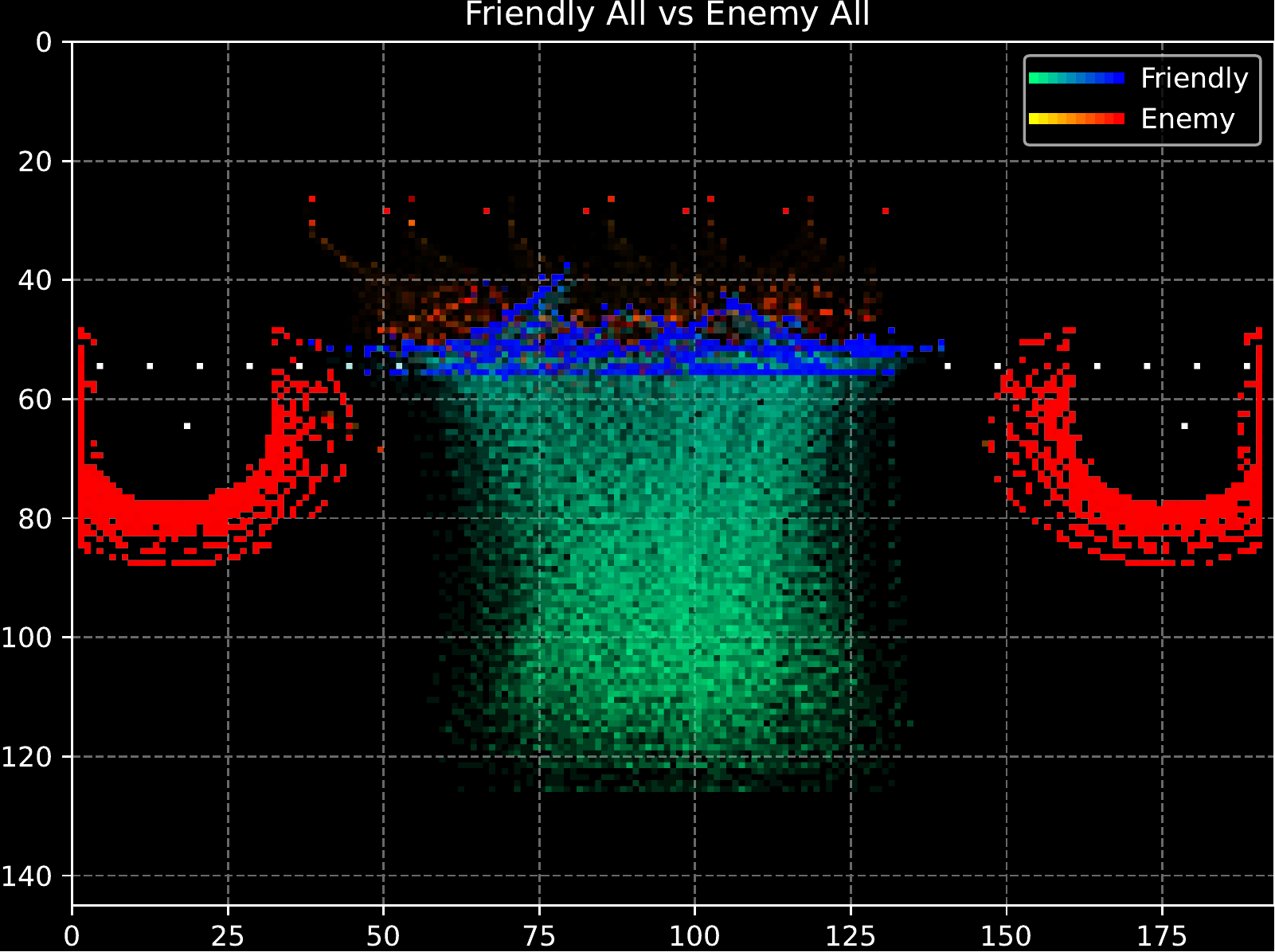}
        \caption{Cluster 2}%
        \label{Fig:Expert2}
    \end{subfigure}%
    \hspace{10pt}
    \begin{subfigure}[b]{0.4\columnwidth}
        \includegraphics[width=\textwidth]{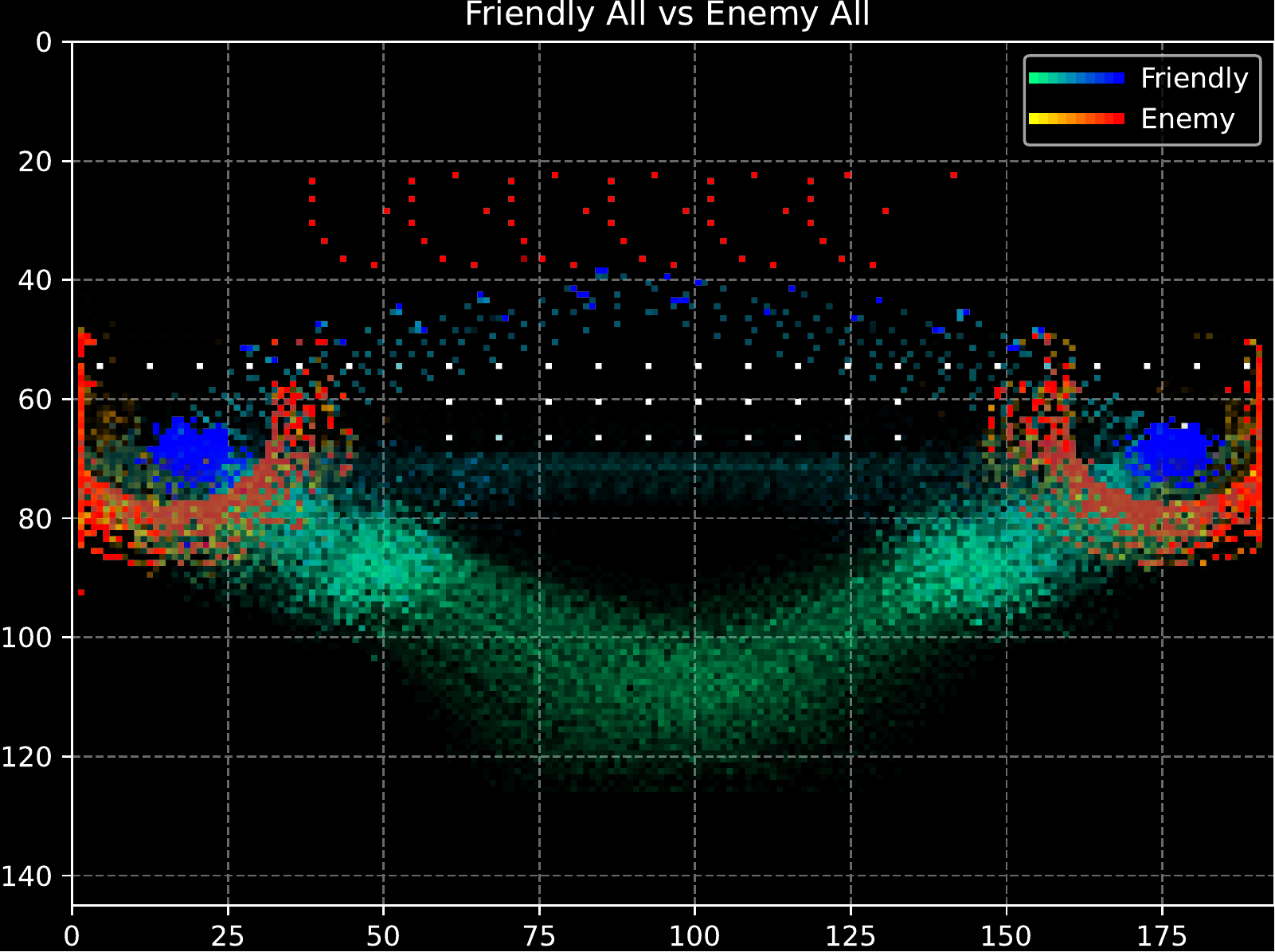}
        \caption{Cluster 3}%
        \label{Fig:Expert3}
    \end{subfigure}
	\caption{Spatio-temporal visualizations of the expert agent's traces for the different clusters. Friendly unit locations are represented with a green-to-blue color palette, while Enemy locations are represented with a yellow-to-red palette. Lighter colors (green, yellow) represent locations occupied earlier in the episode. See text for more details.}%
	\label{Fig:ExpertClusterVis}
\end{figure}
% ......................................................................................

% ......................................................................................
\begin{figure}[!tb]
	\centering
    \begin{subfigure}[b]{0.4\columnwidth}
        \includegraphics[width=\textwidth]{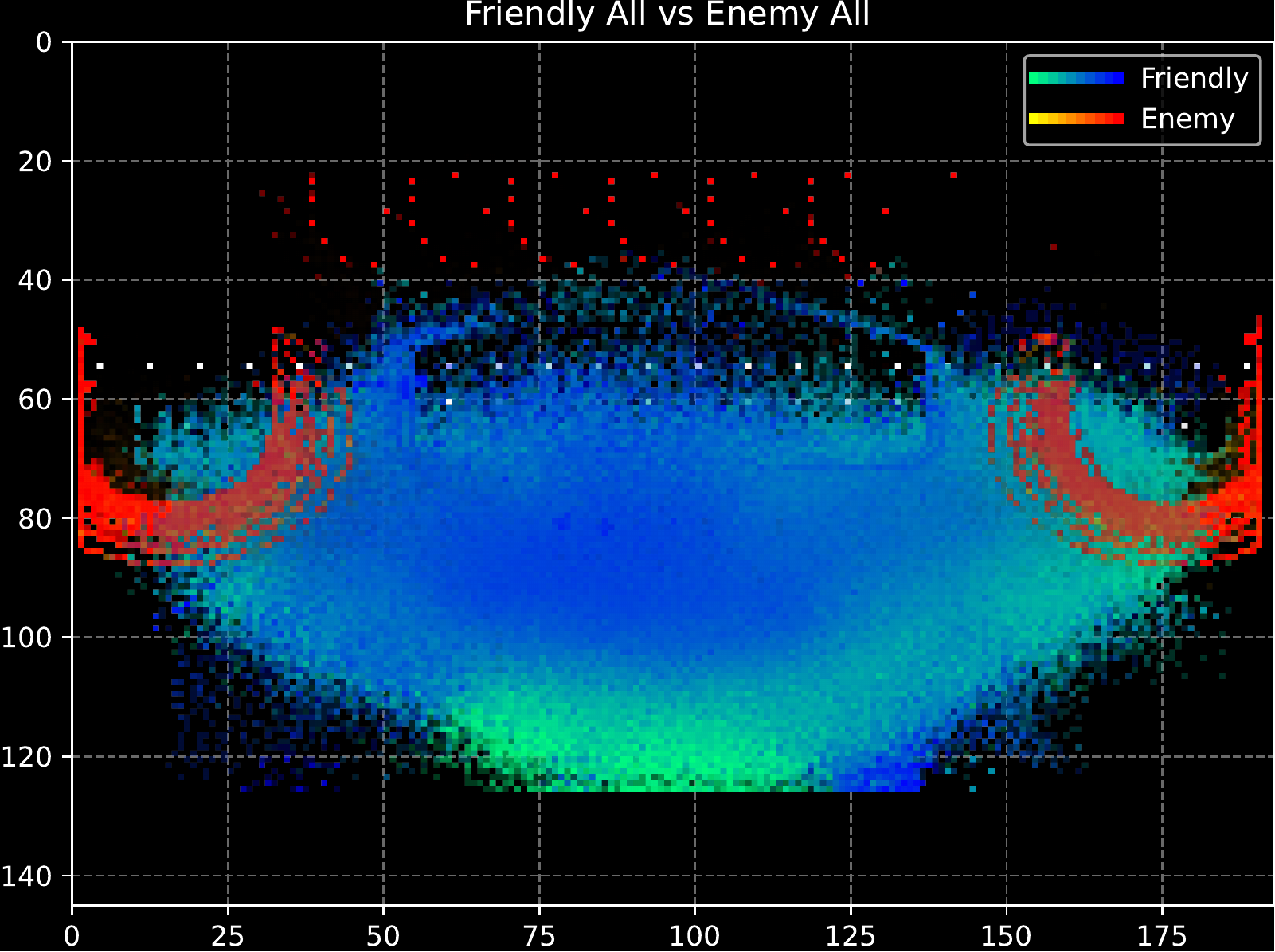}
        \caption{Cluster 0}%
        \label{Fig:RL0}
    \end{subfigure}%
    \hspace{10pt}
    \begin{subfigure}[b]{0.4\columnwidth}
        \includegraphics[width=\textwidth]{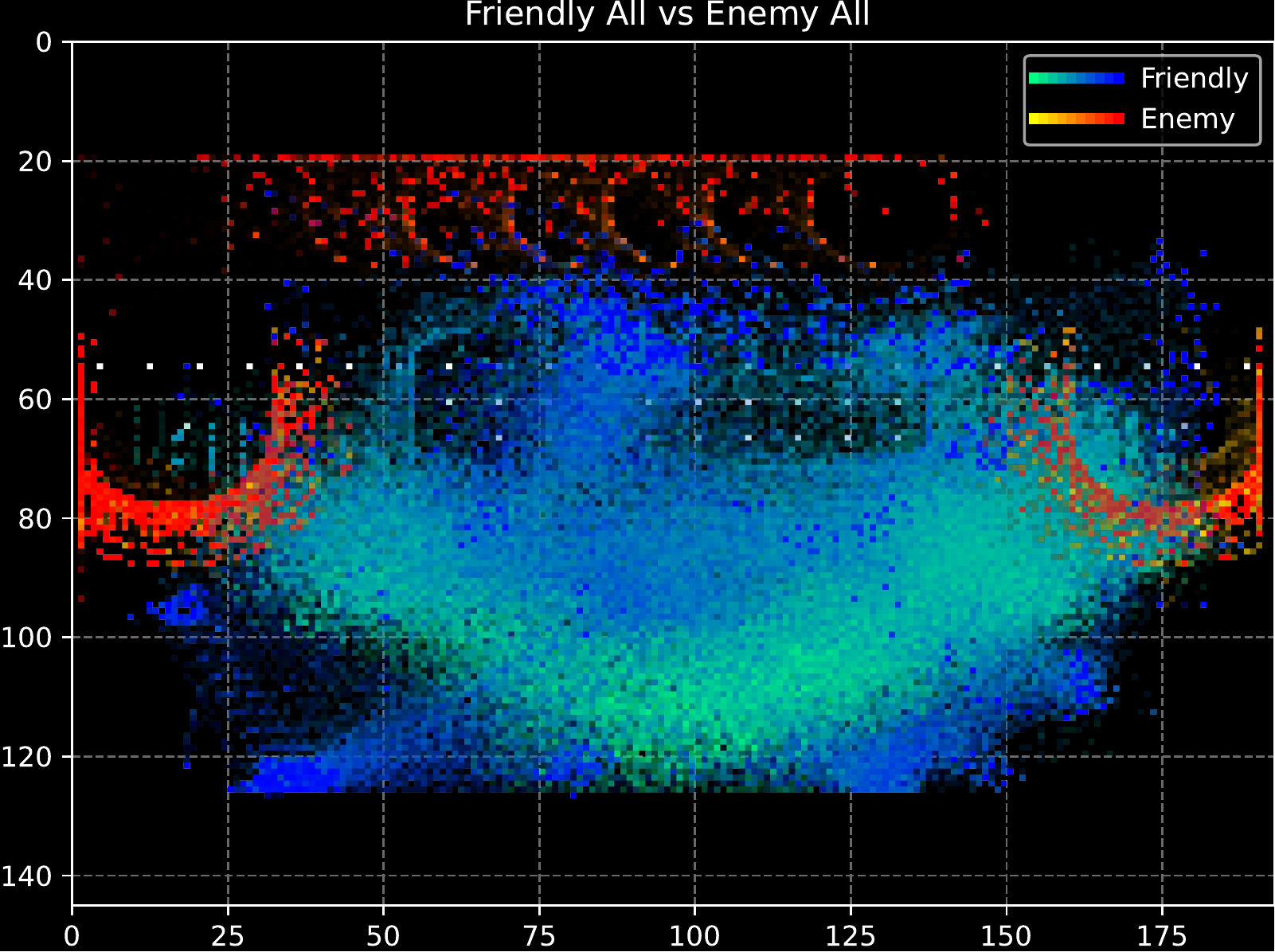}
        \caption{Cluster 1}%
        \label{Fig:RL1}
    \end{subfigure}
    \begin{subfigure}[b]{0.4\columnwidth}
        \includegraphics[width=\textwidth]{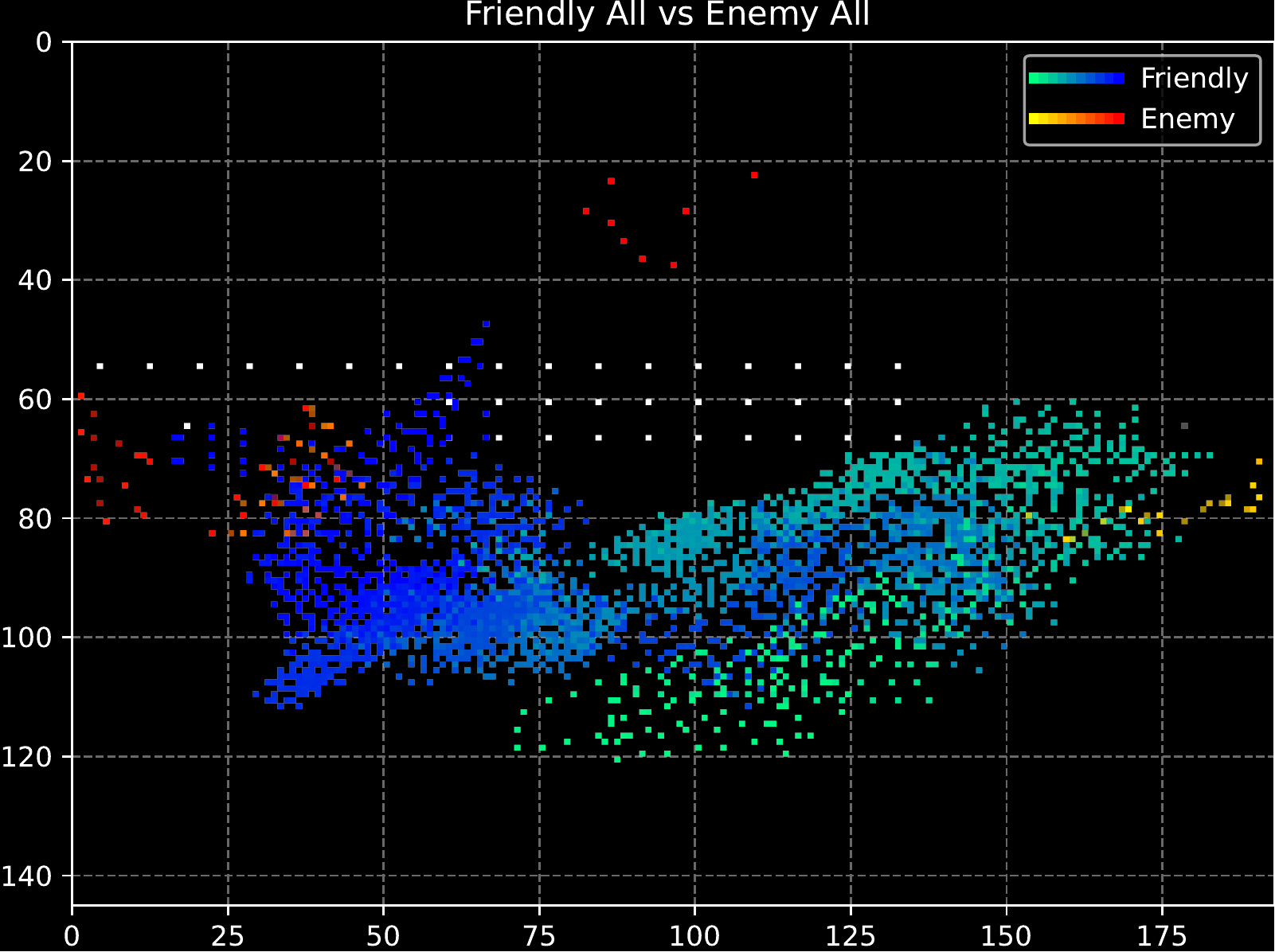}
        \caption{Cluster 2}%
        \label{Fig:RL2}
    \end{subfigure}%
    \hspace{10pt}
    \begin{subfigure}[b]{0.4\columnwidth}
        \includegraphics[width=\textwidth]{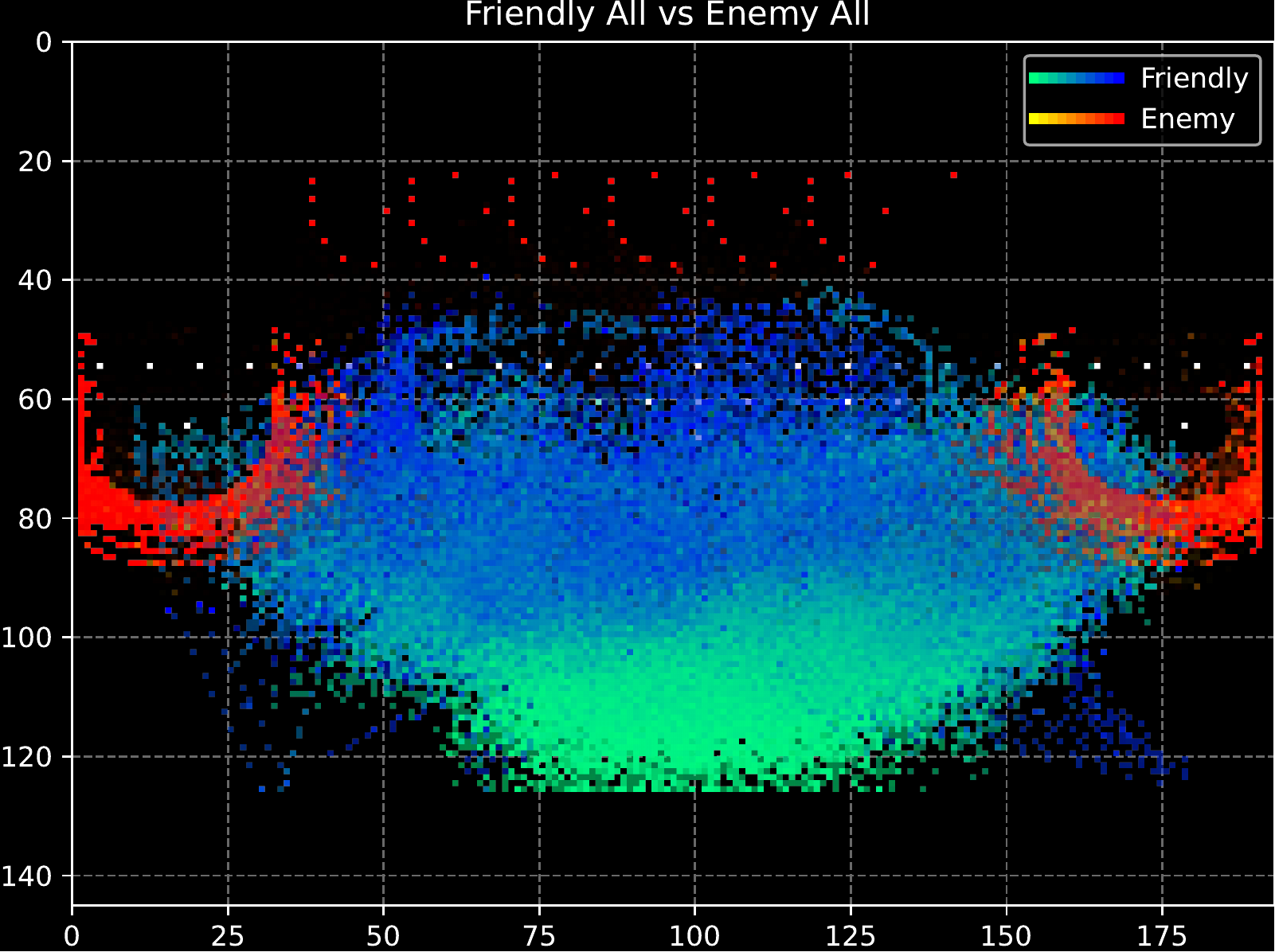}
        \caption{Cluster 3}%
        \label{Fig:RL3}
    \end{subfigure}
	\caption{Spatio-temporal visualizations of the RL agent's traces for the different clusters.}%
	\label{Fig:RLClusterVis}
\end{figure}
% ......................................................................................

% ......................................................................................
\begin{figure}
    \centering
    \includegraphics[width=0.4\textwidth]{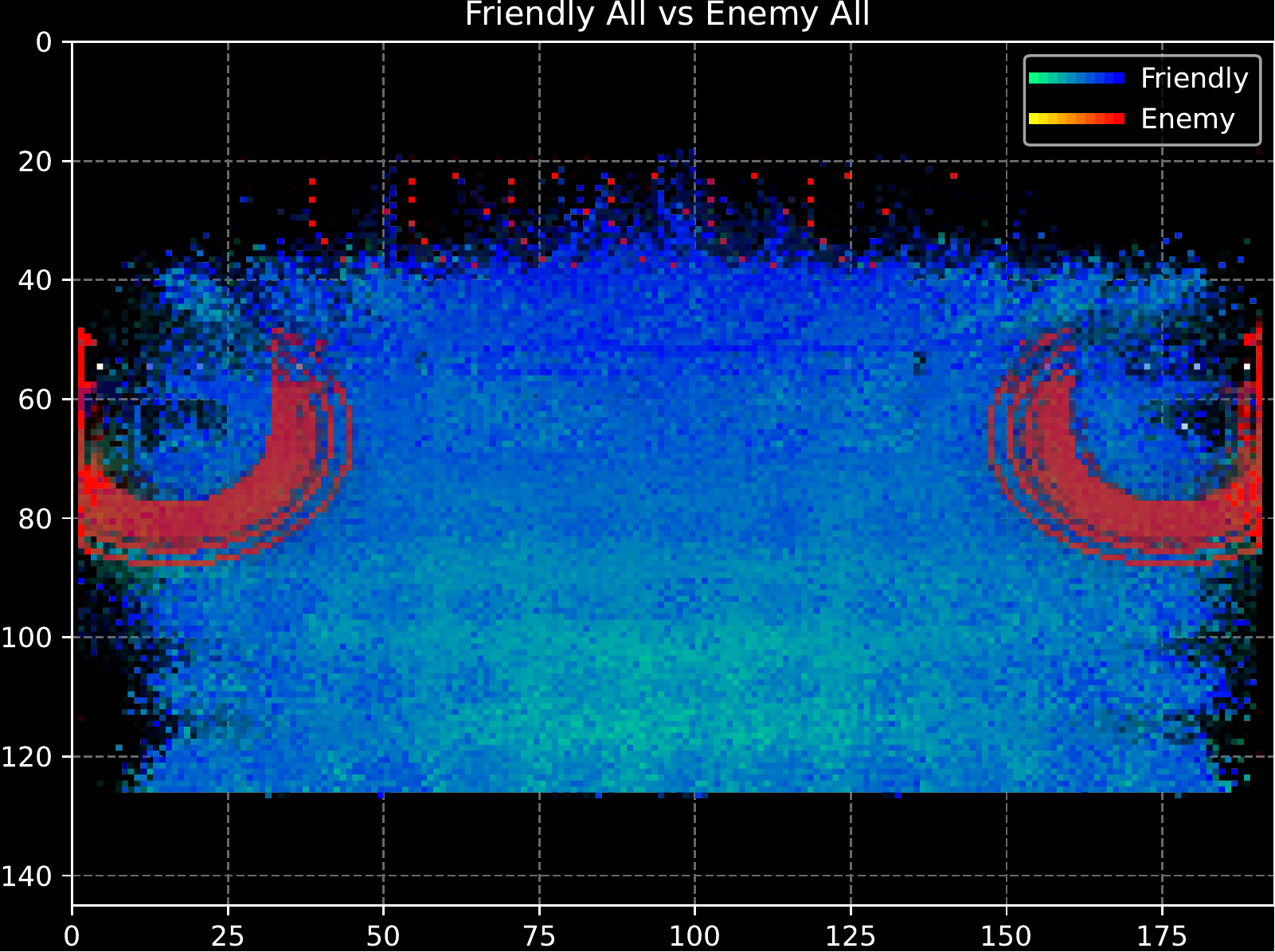}
    \caption{Spatio-temporal visualization of the random agent's traces.}%
	\label{Fig:RandomVis}
\end{figure}
% ......................................................................................

% description of the visualizations
Figs.~\ref{Fig:ExpertClusterVis} and \ref{Fig:RLClusterVis} respectively present the visualizations for the expert and RL agents' clusters. For comparison, we show the visualization of all the random agent traces in Fig.~\ref{Fig:RandomVis}.%
\footnote{The animated versions of the visualizations presented in this paper, as well as videos showing the behavior of the different agents in the task are included as supporting material.}
The visualizations were produced by setting $t=1$, \ie depicting the agents' spatio-temporal behavioral patterns throughout the whole traces. The first thing to notice is that the plots for the expert and RL agents show different behavior patterns whereas the random agent plot in Fig.~\ref{Fig:RandomVis} shows the game board occupied more uniformly. Comparing the expert and the RL agent's plots, we see that the movements for the expert are much more consistent, and that each cluster exhibits distinct patterns corresponding to different strategies and game scenarios. In contrast, the 
%RL agent's visualizations 
visualizations of the RL agent's behavior show that the Friendlies occupy a larger game area in each cluster although, in general, they exhibit an upward movement towards the different objectives in the task. 

% qualitative description of cluster behaviors, expert agent
By looking at the replays of the agents' traces belonging to each cluster, we can qualitatively assess their behavior and the characteristics of the underlying scenarios. Specifically, for the expert agent, cluster $0$ can be characterized by scenarios in which the path to the CC is obstructed and where the \emph{Starport} is absent. In this cluster, the Friendlies frequently destroy at least one secondary objective (seen by the lateral upwards movement in Fig.~\ref{Fig:Expert0}), but the usual outcome is that all the agent's units are destroyed at the end. The expert's behavior in cluster $1$ can be characterized as being cautious---its units either remain still in their initial locations or capture a secondary objective and then remain still until the episode ends. This can be seen by the blue regions at the bottom of the plot and at the secondary objectives locations in Fig.~\ref{Fig:Expert1}. Cluster $2$ captured the scenarios where the path to the CC is unobstructed and where the Friendly force always advances towards its location, as can be seen by the clear upwards movement at the center of the plot in Fig~\ref{Fig:Expert2}. Finally, cluster $3$ captured a complex sequential strategy in which the agent first captures the \emph{Starport}, then gets the \emph{Banshees} (aerial units), then attacks and ends up destroying the CC, while the ground units remain still. Fig.~\ref{Fig:Expert3} captures this pattern by showing faint blue lines going from secondary objective locations towards the CC location on top (the \emph{Banshees}' movements), while the blue regions on each side represent the ground units.

% RL agent
As for the RL agent, cluster $0$ captured situations where its units do not engage the Enemy. In particular, the Friendlies usually try to capture one of the secondary objectives and then just wander around in the center region of the environment while seemingly trying to avoid contact with the Enemy. By looking at the visualizations for the RL agent, Fig.~\ref{Fig:RL0} does show a pattern where the Friendlies occupy the center region at the end of episodes (blue region). Cluster $1$ captured a sequencing strategy similar to that of the expert's in which the Friendlies first capture the \emph{Starport}, and then attack and destroy the CC using the \emph{Banshees}. An interesting tactic we observe with the RL agent is that it often tries to get its ground units away from the Enemy as soon as the \emph{Starport} is destroyed, which is captured in the visualization in Fig.~\ref{Fig:RL1} where we see blue traces going to the bottom of the game board. Cluster $2$ contains only $2$ traces which is insufficient for characterizing a common agent behavior. As for cluster $3$, it represents situations in which the Friendlies are defeated by losing all of its units while trying to destroy one of the secondary objectives. Looking at the corresponding visualization in Fig.~\ref{Fig:RL3}, we see that at the end of episodes the Friendlies seem to be more dispersed that in other clusters, which might be the result of less consistent group attacks.

Overall, these results show that our clustering procedure, based on the high-level feature extractor, partitioned the agents' traces into sets exhibiting meaningfully different behavioral patterns. In addition, they show that our visualization tool allows for capturing the different spatio-temporal patterns resulting from the distinct strategies in each cluster.

% ------------------------------------------------------------------------
\subsection{Strategy Inference}

% general info about the tables
Tables~\ref{Table:ExpertTactics} and \ref{Table:RLTactics} present the best strategies discovered by Strategy Inference for the expert and RL agent's trace clusters, respectively. For each cluster, we first select the top $3$ condition features%
\footnote{In our experiments, this amount of tactics sets seemed sufficient to faithfully characterize the agents' behavior in each cluster. The complete results of the strategy inference component are included as supporting material.}
attaining the highest satisfaction rate in the agent's traces of that cluster compared to the random agent, \ie as measured by $D_{KL}$ (this is presented in the first row of each tactics set). For each feature $f$, we then present the action feature $A_G$ of the \emph{action-goal formula} (defined in Sec.~\ref{Sec:StrategyDiscovery}) that attained the highest $D_{KL}$ score where the goal $G$ matches $f$. Similarly, we also present the action feature $A_C$ of the \emph{condition-action formula} that attained the highest $D_{KL}$ score where the condition $C=f$. Entries marked with ``-'' correspond to situations in which no action formulas were found with $C/G$ matching $f$.%
\footnote{This is expected since some features correspond to the CC being destroyed, which means they cannot be conditions in the \emph{condition-action} formulas. Similarly, some features correspond to the presence of secondary objectives which are defined initially for each scenario and are thus not controllable by the agent, meaning they cannot participate in \emph{action-goal} formulas.}

% ------------------------------------------------------------------------
\subsubsection{Expert Agent Results}

% ......................................................................................
\begin{table}[!htp]
	\centering
	\footnotesize
	\vspace{-20pt}
	\caption{Strategy inference results for the expert agent. We present the $3$ best sets of tactics for each cluster, sorted by $\boldsymbol{D_{KL}}$ score. $\boldsymbol{p}$: formula satisfaction rate for agent's policy. $\boldsymbol{q}$: random agent's satisfaction rate. For each feature $f$ (first row of each tactic set), $A_G$: action feature with highest $D_{KL}$ matching an \emph{action-goal formula} where $G=f$; $A_G$: action with highest $D_{KL}$ matching a \emph{condition-action} formula where $C=f$. $d$ and $r$: matched formula parameters.}%
	\label{Table:ExpertTactics}
	\vspace{5pt}
	\begin{tabular}{l l r r r l }
	\toprule
	\textbf{Cluster} & \textbf{Param.} & $\boldsymbol{p}$ & $\boldsymbol{q}$ & $\boldsymbol{D_{KL}}$ & \textbf{Feature}\\
	\hline
	{\multirow{9}{*}{0}} & $f$ & {$0.40$} & {$0.04$} & {$0.69$} & {\texttt{NOT (Present\_Enemy\_CommandCenter)}}\\
	 & $A_G$ & {$0.39$} & {$0.01$} & {$1.34$} & {\texttt{Target\_Ground\_CommandCenter}, $r=0.9$}\\
	 & $A_C$ &  &  &  & {-}\\
	 \cdashline{2-6}[.5pt/1pt]
	 & $f$ & {$0.22$} & {$0.03$} & {$0.24$} & {\texttt{RelativeCost\_Blue\_Red=advantage}}\\
	 & $A_G$ & {$0.18$} & {$0.01$} & {$0.50$} & {\texttt{Target\_Ground\_CommandCenter}, $r=1.0$}\\
	 & $A_C$ & {$0.02$} & {$0.00$} & {$0.29$} & {\texttt{Target\_Ground\_CommandCenter}, $d=50$, $r=1.0$}\\
	 \cdashline{2-6}[.5pt/1pt]
	 & $f$ & {$0.21$} & {$0.03$} & {$0.22$} & {\texttt{RelativeCost\_Blue\_Red=balanced}}\\
	 & $A_G$ & {$0.02$} & {$0.00$} & {$0.45$} & {\texttt{Target\_Ground\_SiegeTankSieged}, $r=0.7$}\\
	 & $A_C$ & {$0.03$} & {$0.00$} & {$0.53$} & {\texttt{Target\_Ground\_Hellbat}, $d=5$, $r=0.7$}\\
% 	 \cdashline{2-6}[.5pt/1pt]
% 	 & $f$ & {$0.49$} & {$0.22$} & {$0.17$} & {\texttt{Distance\_Blue\_ProductionFacility=melee}}\\
% 	 & $A_G$ & {$0.48$} & {$0.12$} & {$0.40$} & {\texttt{AttackMove\_Friendly\_Ground}, $r=1.0$}\\
% 	 & $A_C$ & {$0.06$} & {$0.00$} & {$1.18$} & {\texttt{AttackMove\_Friendly\_Blue}, $d=200$, $r=1.0$}\\
% 	 \cdashline{2-6}[.5pt/1pt]
% 	 & $f$ & {$0.69$} & {$0.46$} & {$0.11$} & {\texttt{NOT (Present\_Enemy\_Hellion)}}\\
% 	 & $A_G$ & {$0.31$} & {$0.07$} & {$0.25$} & {\texttt{AttackMove\_Friendly\_Ground}, $r=1.0$}\\
% 	 & $A_C$ & {$0.05$} & {$0.00$} & {$0.93$} & {\texttt{Target\_Ground\_CommandCenter}, $d=90$, $r=1.0$}\\
	\midrule
	{\multirow{9}{*}{1}} & $f$ & {$0.72$} & {$0.54$} & {$0.07$} & {\texttt{NOT (Present\_Enemy\_Starport)}}\\
	 & $A_G$ &  &  &  & {-}\\
	 & $A_C$ & {$0.44$} & {$0.00$} & {$1.76$} & {\texttt{NoOp\_Friendly\_Ground}, $d=200$, $r=1.0$}\\
	 \cdashline{2-6}[.5pt/1pt]
	 & $f$ & {$0.74$} & {$0.57$} & {$0.06$} & {\texttt{NOT (Defender\_Starport\_Red)}}\\
	 & $A_G$ &  &  &  & {-}\\
	 & $A_C$ & {$0.45$} & {$0.00$} & {$1.78$} & {\texttt{NoOp\_Friendly\_Ground}, $d=200$, $r=1.0$}\\
	 \cdashline{2-6}[.5pt/1pt]
	 & $f$ & {$0.66$} & {$0.49$} & {$0.06$} & {\texttt{Defender\_CommandCenter\_Red}}\\
	 & $A_G$ &  &  &  & {-}\\
	 & $A_C$ & {$0.34$} & {$0.00$} & {$7.27$} & {\texttt{NoOp\_Friendly\_Ground}, $d=200$, $r=1.0$}\\
% 	 \cdashline{2-6}[.5pt/1pt]
% 	 & $f$ & {$0.66$} & {$0.49$} & {$0.06$} & {\texttt{Present\_Enemy\_SiegeTankSieged}}\\
% 	 & $A_G$ &  &  &  & {-}\\
% 	 & $A_C$ & {$0.34$} & {$0.00$} & {$7.27$} & {\texttt{NoOp\_Friendly\_Ground}, $d=200$, $r=1.0$}\\
% 	 \cdashline{2-6}[.5pt/1pt]
% 	 & $f$ & {$0.63$} & {$0.49$} & {$0.04$} & {\texttt{Defender\_EngineeringBay\_Red}}\\
% 	 & $A_G$ &  &  &  & {-}\\
% 	 & $A_C$ & {$0.33$} & {$0.00$} & {$1.31$} & {\texttt{NoOp\_Friendly\_Ground}, $d=200$, $r=1.0$}\\
	\midrule
	{\multirow{9}{*}{2}} & $f$ & {$0.72$} & {$0.04$} & {$1.83$} & {\texttt{NOT (Present\_Enemy\_CommandCenter)}}\\
	 & $A_G$ & {$0.71$} & {$0.01$} & {$3.06$} & {\texttt{Target\_Ground\_CommandCenter}, $r=0.9$}\\
	 & $A_C$ &  &  &  & {-}\\
	 \cdashline{2-6}[.5pt/1pt]
	 & $f$ & {$1.00$} & {$0.50$} & {$0.70$} & {\texttt{Distance\_Blue\_CommandCenter=close}}\\
	 & $A_G$ & {$1.00$} & {$0.18$} & {$1.74$} & {\texttt{Target\_Ground\_CommandCenter}, $r=1.0$}\\
	 & $A_C$ & {$0.75$} & {$0.00$} & {$3.58$} & {\texttt{Target\_Ground\_CommandCenter}, $d=50$, $r=1.0$}\\
	 \cdashline{2-6}[.5pt/1pt]
	 & $f$ & {$1.00$} & {$0.52$} & {$0.65$} & {\texttt{NOT (Defender\_CommandCenter\_Red)}}\\
	 & $A_G$ &  &  &  & {-}\\
	 & $A_C$ & {$0.78$} & {$0.01$} & {$3.13$} & {\texttt{Target\_Ground\_CommandCenter}, $d=60$, $r=1.0$}\\
% 	 \cdashline{2-6}[.5pt/1pt]
% 	 & $f$ & {$1.00$} & {$0.52$} & {$0.65$} & {\texttt{NOT (Present\_Enemy\_SiegeTankSieged)}}\\
% 	 & $A_G$ &  &  &  & {-}\\
% 	 & $A_C$ & {$0.78$} & {$0.01$} & {$3.13$} & {\texttt{Target\_Ground\_CommandCenter}, $d=60$, $r=1.0$}\\
% 	 \cdashline{2-6}[.5pt/1pt]
% 	 & $f$ & {$0.25$} & {$0.13$} & {$0.06$} & {\texttt{NOT (Present\_Enemy\_Obstacle)}}\\
% 	 & $A_G$ &  &  &  & {-}\\
% 	 & $A_C$ & {$0.15$} & {$0.00$} & {$2.97$} & {\texttt{Target\_Ground\_CommandCenter}, $d=70$, $r=1.0$}\\
	\midrule
	{\multirow{9}{*}{3}} & $f$ & {$1.00$} & {$0.04$} & {$3.35$} & {\texttt{NOT (Present\_Enemy\_CommandCenter)}}\\
	 & $A_G$ & {$1.00$} & {$0.00$} & {$23.03$} & {\texttt{Target\_Air\_CommandCenter}, $r=0.7$}\\
	 & $A_C$ &  &  &  & {-}\\
	 \cdashline{2-6}[.5pt/1pt]
	 & $f$ & {$1.00$} & {$0.05$} & {$3.06$} & {\texttt{Present\_Friendly\_Air}}\\
	 & $A_G$ & {$1.00$} & {$0.03$} & {$3.65$} & {\texttt{NoOp\_Friendly\_Ground}, $r=1.0$}\\
	 & $A_C$ & {$1.00$} & {$0.00$} & {$23.03$} & {\texttt{Target\_Air\_CommandCenter}, $d=60.0$, $r=1.0$}\\
	 \cdashline{2-6}[.5pt/1pt]
	 & $f$ & {$1.00$} & {$0.50$} & {$0.70$} & {\texttt{Distance\_Blue\_CommandCenter=close}}\\
	 & $A_G$ & {$1.00$} & {$0.01$} & {$5.12$} & {\texttt{Target\_Air\_CommandCenter}, $r=0.7$}\\
	 & $A_C$ & {$1.00$} & {$0.00$} & {$23.03$} & {\texttt{Target\_Air\_CommandCenter}, $d=50$, $r=1.0$}\\
% 	 \cdashline{2-6}[.5pt/1pt]
% 	 & $f$ & {$1.00$} & {$0.51$} & {$0.68$} & {\texttt{Defender\_Starport\_Red}}\\
% 	 & $A_G$ &  &  &  & {-}\\
% 	 & $A_C$ & {$1.00$} & {$0.00$} & {$23.03$} & {\texttt{Target\_Air\_CommandCenter}, $d=60$, $r=0.7$}\\
% 	 \cdashline{2-6}[.5pt/1pt]
% 	 & $f$ & {$1.00$} & {$0.51$} & {$0.68$} & {\texttt{Present\_Enemy\_Starport}}\\
% 	 & $A_G$ &  &  &  & {-}\\
% 	 & $A_C$ & {$1.00$} & {$0.00$} & {$23.03$} & {\texttt{Target\_Air\_CommandCenter}, $d=40$, $r=0.9$}\\
	\bottomrule
	\end{tabular}
\end{table}
% ......................................................................................

% Expert results
By looking at the results of the expert agent in Table~\ref{Table:ExpertTactics}, we see a good match between the discovered tactics and the hand-coded behavior rules listed in Sec.~\ref{Sec:ExpertDef}. Further, the strategies reinforce the empirical analysis of the agent behavior in each cluster presented in the previous section. In particular, for cluster $0$, the main goal was identified as destroying the CC (\texttt{NOT (Present\_Enemy\_CommandCenter)}). Also, Strategy Inference captured tactics that target the CC, its defenders and reinforcements using only ground units, which can explain the relative poor performance in this cluster's traces. Further, recall that the relative strength between the forces (measured through the combined cost of their units) plays a key role in the expert agent's strategy, as defined by the hand-coded rules. The tactics discovered for this cluster were able to recover that aspect of the expert's behavior, where the relative cost (\texttt{RelativeCost\_Blue\_Red=*}) conditions determine who the target of the action is: the CC, if in advantage, the CC's defenders and the reinforcements, if balanced. %Interestingly, the last tactics group corresponds to the tactic of attacking with the ground units until a secondary objective is destroyed (resulting in the absence of Hellion units), \emph{before} attacking the CC. 

Similarly, the tactics discovered for cluster $1$ captured the cautious nature of the expert's behavior that are inherent to the behavior rules, which is denoted by the \texttt{NoOp} action tactics. Further, Strategy Inference captured the features on which such behavior is conditioned, namely the absence of the \emph{Starport} (and consequentially of the aerial units) and the presence of CC defenders. 

For cluster $2$, as observed from its spatio-temporal visualization, there are no obstacle units between the agent's units and the CC, so the hand-coded strategy is to attack the CC directly. This is consistent with the best goal tactics discovered for this cluster, where the agent attacks the CC with the ground units until they are destroyed (\texttt{NOT (Present\_Enemy\_CommandCenter)}). Further, one set of tactics identified the absence of CC defenders (\texttt{NOT (Defender\_CommandCenter\_Red)}) as a condition for attacking the CC, which again is consistent with the rules. 
%One of the discovered condition features was \texttt{NOT (Present\_Enemy\_Obstacle)}, which matches exactly the conditions of the scenarios in this cluster. 

The tactics discovered for cluster $3$ are also consistent with the behavior resulting from the hand-coded strategy: they capture the tactic of attacking the CC using the aerial units. Interestingly, one set also captured the fact that the ground units do not engage the enemy (\texttt{NoOp\_Friendly\_Ground}) while the agent's aerial units are attacking the CC.

% ------------------------------------------------------------------------
\subsubsection{RL Agent Results}

% ......................................................................................
\begin{table}[!htp]
	\centering
	\footnotesize
	\caption{Strategy inference results for the RL agent. See text for more details.}%
	\label{Table:RLTactics}
	\begin{tabular}{l l r r r l }
	\toprule
	\textbf{Cluster} & \textbf{Param.} & $\boldsymbol{p}$ & $\boldsymbol{q}$ & $\boldsymbol{D_{KL}}$ & \textbf{Feature}\\
	\hline
	{\multirow{9}{*}{0}} & {$f$} & {$0.79$} & {$0.46$} & {$0.23$} & {\texttt{NOT (Present\_Enemy\_Hellion)}}\\
	 & {$A_G$} & {$0.05$} & {$0.00$} & {$1.00$} & {\texttt{Retreating\_Friendly\_Blue\_Red}, $r=0.8$}\\
	 & {$A_C$} & {$0.79$} & {$0.24$} & {$0.67$} & {\texttt{Target\_Ground\_CommandCenter}, $d=0$, $r=0.7$}\\
	 \cdashline{2-6}[.5pt/1pt]
	 & {$f$} & {$0.96$} & {$0.75$} & {$0.17$} & {\texttt{Distance\_Blue\_ProductionFacility=close}}\\
	 & {$A_G$} & {$0.53$} & {$0.14$} & {$0.42$} & {\texttt{Advancing\_Friendly\_Ground\_ProductionFacility}, $r=1.0$}\\
	 & {$A_C$} & {$0.69$} & {$0.09$} & {$1.07$} & {\texttt{AttackMove\_Friendly\_Blue}, $d=200$, $r=0.7$}\\
	 \cdashline{2-6}[.5pt/1pt]
	 & {$f$} & {$0.70$} & {$0.46$} & {$0.12$} & {\texttt{NOT (Present\_Enemy\_Marauder)}}\\
	 & {$A_G$} & {$0.01$} & {$0.00$} & {$0.27$} & {\texttt{Advancing\_Friendly\_Blue\_CommandCenter}, $r=0.7$}\\
	 & {$A_C$} & {$0.70$} & {$0.24$} & {$0.49$} & {\texttt{Target\_Ground\_CommandCenter}, $d=0$, $r=0.7$}\\
% 	 \cdashline{2-6}[.5pt/1pt]
% 	 & {$f$} & {$0.40$} & {$0.22$} & {$0.08$} & {\texttt{Distance\_Blue\_ProductionFacility=melee}}\\
% 	 & {$A_G$} & {$0.36$} & {$0.12$} & {$0.21$} & {\texttt{AttackMove\_Friendly\_Ground}, $r=1.0$}\\
% 	 & {$A_C$} & {$0.19$} & {$0.02$} & {$0.30$} & {\texttt{AttackMove\_Friendly\_Blue}, $d=200$, $r=0.7$}\\
% 	 \cdashline{2-6}[.5pt/1pt]
% 	 & {$f$} & {$0.70$} & {$0.54$} & {$0.05$} & {\texttt{NOT (Defender\_Factory\_Red)}}\\
% 	 & {$A_G$} & {$0.23$} & {$0.03$} & {$0.28$} & {\texttt{AttackMove\_Friendly\_Ground}, $r=1.0$}\\
% 	 & {$A_C$} & {$0.51$} & {$0.14$} & {$0.39$} & {\texttt{AttackMove\_Friendly\_Blue}, $d=200$, $r=0.7$}\\
	\midrule
	{\multirow{9}{*}{1}} & {$f$} & {$1.00$} & {$0.05$} & {$3.06$} & {\texttt{Present\_Friendly\_Air}}\\
	 & {$A_G$} & {$0.99$} & {$0.03$} & {$3.38$} & {\texttt{AttackMove\_Friendly\_Blue}, $r=0.9$}\\
	 & {$A_C$} & {$0.39$} & {$0.00$} & {$8.22$} & {\texttt{AttackMove\_Friendly\_Air}, $d=100$, $r=0.9$}\\
	 \cdashline{2-6}[.5pt/1pt]
	 & {$f$} & {$0.70$} & {$0.04$} & {$1.76$} & {\texttt{NOT (Present\_Enemy\_CommandCenter)}}\\
	 & {$A_G$} & {$0.54$} & {$0.00$} & {$3.07$} & {\texttt{AttackMove\_Friendly\_Air}, $r=0.7$}\\
	 & {$A_C$} &  &  &  & {-}\\
	 \cdashline{2-6}[.5pt/1pt]
	 & {$f$} & {$1.00$} & {$0.51$} & {$0.68$} & {\texttt{Defender\_Starport\_Red}}\\
	 & {$A_G$} &  &  &  & {-}\\
	 & {$A_C$} & {$0.49$} & {$0.00$} & {$10.53$} & {\texttt{AttackMove\_Friendly\_Air}, $d=100$, $r=0.7$}\\
% 	 \cdashline{2-6}[.5pt/1pt]
% 	 & {$f$} & {$1.00$} & {$0.51$} & {$0.68$} & {\texttt{Present\_Enemy\_Starport}}\\
% 	 & {$A_G$} &  &  &  & {-}\\
% 	 & {$A_C$} & {$0.52$} & {$0.00$} & {$11.26$} & {\texttt{AttackMove\_Friendly\_Air}, $d=100$, $r=0.7$}\\
% 	 \cdashline{2-6}[.5pt/1pt]
% 	 & {$f$} & {$0.38$} & {$0.03$} & {$0.64$} & {\texttt{RelativeCost\_Blue\_Red=advantage}}\\
% 	 & {$A_G$} & {$0.04$} & {$0.00$} & {$0.71$} & {\texttt{Advancing\_Friendly\_Ground\_Mobile}, $r=0.8$}\\
% 	 & {$A_C$} &  &  &  & {-}\\
	\midrule
	{\multirow{9}{*}{2}} & {$f$} & {$1.00$} & {$0.63$} & {$0.43$} & {\texttt{NOT (Present\_Friendly\_Blue)}}\\
	 & {$A_G$} & {$0.96$} & {$0.44$} & {$0.66$} & {\texttt{AttackMove\_Friendly\_Ground}, $r=0.7$}\\
	 & {$A_C$} &  &  &  & {-}\\
	 \cdashline{2-6}[.5pt/1pt]
	 & {$f$} & {$1.00$} & {$0.85$} & {$0.17$} & {\texttt{Between\_Red\_Blue\_CommandCenter}}\\
	 & {$A_G$} & {$0.99$} & {$0.63$} & {$0.41$} & {\texttt{AttackMove\_Friendly\_Ground}, $r=0.7$}\\
	 & {$A_C$} & {$0.86$} & {$0.29$} & {$0.70$} & {\texttt{Target\_Ground\_CommandCenter}, $d=0$, $r=0.7$}\\
	 \cdashline{2-6}[.5pt/1pt]
	 & {$f$} & {$1.00$} & {$0.93$} & {$0.06$} & {\texttt{Between\_Red\_Blue\_ProductionFacility}}\\
	 & {$A_G$} & {$0.93$} & {$0.69$} & {$0.17$} & {\texttt{AttackMove\_Friendly\_Ground}, $r=0.7$}\\
	 & {$A_C$} & {$0.85$} & {$0.37$} & {$0.50$} & {\texttt{Target\_Ground\_CommandCenter}, $d=0$, $r=0.7$}\\
% 	 \cdashline{2-6}[.5pt/1pt]
% 	 & {$f$} & {$0.82$} & {$0.68$} & {$0.05$} & {\texttt{Present\_Enemy\_Marauder}}\\
% 	 & {$A_G$} &  &  &  & {-}\\
% 	 & {$A_C$} & {$0.82$} & {$0.34$} & {$0.50$} & {\texttt{Target\_Ground\_Marauder}, $d=0$, $r=0.7$}\\
% 	 \cdashline{2-6}[.5pt/1pt]
% 	 & {$f$} & {$0.60$} & {$0.46$} & {$0.04$} & {\texttt{NOT (Present\_Enemy\_Hellion)}}\\
% 	 & {$A_G$} & {$0.02$} & {$0.00$} & {$0.45$} & {\texttt{Retreating\_Friendly\_Blue\_Red}, $r=0.8$}\\
% 	 & {$A_C$} & {$0.60$} & {$0.24$} & {$0.29$} & {\texttt{Target\_Ground\_CommandCenter}, $d=0$, $r=0.7$}\\
	\midrule
	{\multirow{9}{*}{3}} & {$f$} & {$1.00$} & {$0.01$} & {$4.27$} & {\texttt{NOT (Present\_Enemy\_ProductionFacility)}}\\
	 & {$A_G$} & {$1.00$} & {$0.00$} & {$5.52$} & {\texttt{AttackMove\_Friendly\_Blue}, $r=1.0$}\\
	 & {$A_C$} & {$0.50$} & {$0.00$} & {$10.82$} & {\texttt{Target\_Ground\_SiegeTankSieged}, $d=0$, $r=0.7$}\\
	 \cdashline{2-6}[.5pt/1pt]
	 & {$f$} & {$1.00$} & {$0.03$} & {$3.65$} & {\texttt{NOT (Defender\_ProductionFacility\_Red)}}\\
	 & {$A_G$} & {$1.00$} & {$0.01$} & {$4.61$} & {\texttt{MoveGrid\_Friendly\_Ground}, $r=1.0$}\\
	 & {$A_C$} & {$0.50$} & {$0.00$} & {$10.82$} & {\texttt{Target\_Ground\_SiegeTankSieged}, $d=0$, $r=0.7$}\\
	 \cdashline{2-6}[.5pt/1pt]
	 & {$f$} & {$1.00$} & {$0.03$} & {$3.38$} & {\texttt{RelativeCost\_Blue\_Red=advantage}}\\
	 & {$A_G$} & {$1.00$} & {$0.02$} & {$4.20$} & {\texttt{AttackMove\_Friendly\_Ground}, $r=1.0$}\\
	 & {$A_C$} & {$0.50$} & {$0.00$} & {$10.82$} & {\texttt{Target\_Ground\_SiegeTankSieged}, $d=0$, $r=0.7$}\\
% 	 \cdashline{2-6}[.5pt/1pt]
% 	 & {$f$} & {$1.00$} & {$0.05$} & {$3.06$} & {\texttt{Present\_Friendly\_Air}}\\
% 	 & {$A_G$} & {$1.00$} & {$0.02$} & {$3.86$} & {\texttt{MoveGrid\_Friendly\_Ground}, $r=1.0$}\\
% 	 & {$A_C$} & {$1.00$} & {$0.02$} & {$3.82$} & {\texttt{Target\_Ground\_CommandCenter}, $d=0$, $r=0.7$}\\
% 	 \cdashline{2-6}[.5pt/1pt]
% 	 & {$f$} & {$0.50$} & {$0.03$} & {$1.01$} & {\texttt{RelativeCost\_Blue\_Red=balanced}}\\
% 	 & {$A_G$} & {$0.50$} & {$0.01$} & {$1.48$} & {\texttt{MoveGrid\_Friendly\_Ground}, $r=1.0$}\\
% 	 & {$A_C$} &  &  &  & {-}\\
	\bottomrule
	\end{tabular}
\end{table}
% ......................................................................................

% RL results
The strategy inference results for the RL agent, listed in Table~\ref{Table:RLTactics}, also provide valuable insights about the agent's behavior. Specifically, looking at the tactics discovered for cluster $0$, we see that the resulting strategy captures mainly two behaviors: 1) the second set of tactics captured the behavior of advancing towards a Production Facility (secondary objective) until the units are close to it, followed by attacking the enemy; 2) the first and third set of tactics capture the (subsequent) behavior of attacking the CC when enemy Hellions and Marauders are not present, which occurs when the secondary objectives that they are defending are destroyed (this results in no aerial reinforcements).%
\footnote{The action-goal tactics in these sets ($A_G$) can be discarded since the difference between the formulas' satisfaction rate for the RL agent, $p$, and for the random agent, $q$, is considerably low.}
Further, the results of this cluster seem to indicate that the ``wandering'' behavior of the agent's units observed from the traces' visualization are a result of the agent's indecisiveness in the absence of aerial units rather than some purposeful, learned behavior---which would otherwise have been captured by tactics discovered through Strategy Inference.

As for cluster $1$, overall the results denote the strategy of attacking the Enemy (\texttt{AttackMove\_Friendly\_Blue}) until the aerial units are present (thus destroying the Starport), and then attacking the enemy with the aerial units (\texttt{AttackMove\_Friendly\_Air}) until the CC is no longer present, which is consistent with our observations of the agent's behavior in this cluster. Interestingly, we see that the preferred action by the RL agent is AttackMove instead of targeting the CC, resulting in the same outcome. %In addition, one interesting discovered tactic was that of advancing with the ground units \emph{only when} the relative cost is advantageous for the agent (see last group). 
Regarding cluster $2$, as mentioned earlier it contains only two traces, so we refrain from making any conclusions from the resulting tactics. 

Finally, the discovered tactics for cluster $3$ did capture a behavior consistent with our observations of the agent frequently trying to destroy (the defenders of) a Production Facility (corresponding to achieving a secondary objective). Interestingly, it discovered the tactic of first destroying a secondary objective \emph{and then} approaching the CC (\texttt{Target\_Ground\_SiegeTankSieged} targets the defenders of the CC). %We can also see that even in situations where it captured a \emph{Starport}, thus leading to \texttt{Present\_Friendly\_Air}, it still attacked the CC with ground units, which is highly undesirable since it leads to the loss of some or all of those units. Also, we see that it usually targets the CC defenders instead of the CC directly. 
These results are consistent with the observed outcome of the traces in this cluster, where the agent ended up losing all its units since it is difficult to capture a defended CC without aerial units. 

Together, many of the discovered tactics for the RL agent's clusters help explain why it could not achieve a very high performance, and also that further training would be required, \eg to learn to target the enemy more efficiently. Moreover, compared to the expert agent's results, the discovered strategies are not as consistent, denoting more tactics and less coherent behavior in each cluster, as is expected given the RL agent's performance level in the task.

% RELATED WORK
% =========================================
\section{Related Work}%
\label{Sec:Related}

The large body of work in eXplainable AI (XAI) over the last decade \citep{barredoarrieta2020xai,goebel2018xai} has been driven by the rapid advances in deep learning that have resulted in high-performing but opaque learned models. With respect to the XAI taxonomy in \citep{barredoarrieta2020xai}, our approach is a post-hoc, model-agnostic technique that uses feature-relevance-based inference and derives textual descriptions and visual representations of agent strategies. Much of the work in XAI has been focused on classification tasks and thus on explaining single decisions \ie class predictions. In the sequential decision tasks addressed by reinforcement learning, agents make a series of decisions that affect and are influenced by environmental dynamics. This opens up a number of opportunities for explaining different aspects of the learned policy and has led to a variety of approaches to eXplainable Reinforcement Learning (XRL) over the last few years. Recent surveys \citep{heuillet2021xrl,puiutta2020xrl} provide detailed descriptions of many of the existing XRL methods; here we briefly highlight works that are mostly related to our approach according to the aforementioned taxonomy.%
\footnote{Although our approach has some connections with inverse reinforcement learning, our goal is to find interpretable descriptions for strategies rather than a (possibly opaque) reward function denoting the agent's overall goal.}

Query-based approaches provide explanations for specific situations. For example, the work in \citep{hayes2017policyexplanation} learns a queryable model of an agent policy to let human collaborators analyze what a robot has learned through templated questions around identifying conditions for actions, predicting what an agent will do, and explaining expectation violations. The approach finds correlations between conditions and actions but does not try to explain the motivation (end goal) behind those actions. The query-based approach in \citep{vanderwaa2018contrastive} answers questions about actions and policy in terms of consequences. Like our approach, it transforms raw features into ``higher-level descriptors'' for conditions and outcomes to facilitate explanation, but uses simulation to predict sequences of future actions. In contrast to these approaches, ours finds descriptions for \emph{all} the strategies learned by the agent for different situations and does not require user queries to initiate explanation (although it could easily serve that purpose by returning the tactics matching specific query conditions).

Another approach is to extract examples of behavior denoted by relevant agent trajectories to be presented to a user. \citet{huang2019enabling} select \emph{maximally informative} examples that best describe the reward function of the agent against other reward functions based on algorithmic teaching techniques and different distance functions to select trajectories. A similar approach in \citep{lage2019exploring} produces summaries of policies by extracting trajectories that maximize the \emph{quality} of their reconstruction according to the principle of maximum entropy. \citet{sequeira2020interestingness,sequeira2019interestingness} use the concept of \emph{interestingness} to identify key decision states given (possibly long) agent trajectories, and produce short video clips showing the agent's behavior during those important moments, \ie behavior summaries. Unlike these approaches relying on behavior \emph{examples}, we focus on extracting general strategy descriptions.

More similar to our work are approaches that extract structural models of policies denoting different agent strategies. For example, in \citep{madumal2020causallens}, contrastive (why/why not) explanations are provided about single task predictions, with the explanations derived from structural causal models in the form of action influence models. \citet{koul2018learning} learn finite state representations of deep RL policies that abstract the learned policy into a simpler, more interpretable structure, but the generated representations are hard to interpret since the semantics behind the discretized states are unknown. In \citep{dereszynski2011learning,hostetler2012inferring}, probabilistic graphical model techniques are used to learn finite state models of strategy for the StarCraft domain, but work in this space has tended to focus on strategy at a higher level of abstraction rather than in terms of actual policy actions.

Our approach to strategy inference borrows from \citep{NEURIPS2018_taskspec} the use of KL divergence against a random agent to determine which formulas best characterize an agent's strategy. We build on that work to deal with more realistic scenarios, where we are not given perfect demonstration sets, and expand the scope to deal with much larger and more complex domains. In \cite{AsarinParametric}, Parametric Signal Temporal Logic (PSTL) formulas are synthesized, but the formulas are specified ahead of time up to numeric parameters only, and focus more on characterizing properties of continuous-time and continuous-valued signals. The approach in \cite{DBLP:journals/corr/abs-1806-03953} looks for the minimal formula that is consistent with a set of traces. However, in complex settings such as SC2, such minimal formulas are always trivial due to the large number of invariant environmental conditions. \citet{7371998} mine LTL properties from logs of software behavior. %Similarly to our work, 
As in our work, they make use of formula templates that are instantiated with atomic propositions
%. Also, 
and their notion of confidence serves a similar purpose as the minimum satisfaction rates in our soft temporal operators. However, they do not have a way to distinguish intentional behavior (\ie strategies) from accidentally correlated features. We have found this to be the core of the strategy discovery problem, which we address by using the $D_{KL}$ score against a random policy.

% CONCLUSIONS
% =========================================
\section{Conclusions and Ongoing Work}%
\label{Sec:Conclusions}

In this paper, we presented a framework for extracting descriptions of agent strategies from traces of their behavior. Ours is a post-hoc approach to XRL, requiring only traces of agent behavior as input. It can thus be applied to describe the strategies of any agent, regardless of how the agent's policy is learned or even whether the policy is learned at all. 

The approach involves first partitioning the traces into a set of clusters that capture the different strategies the agent uses to accomplish its task. We use a novel embedding scheme that combines discounted feature sums with sequence graph transforms to find frequent patterns over conditions and action sequences. We then find temporal logic formulas to describe the strategies in the clusters, using an information-theoretic approach to select the most meaningful formulas as the ones that best distinguish the observed behavior from that of a random agent. 

We tested our approach on SC2 force-on-force combat scenarios. We created a high-level feature extractor for SC2 capable of characterizing the state of the environment and the agent's local behavior. Our experiments show that we can successfully use our framework to find clusters of traces that display different behavioral patterns and to find meaningful strategy descriptions that capture the behavior observed in those clusters. To help with understanding the agents' behavior in each cluster and the environment conditions leading to it, we designed a visualization tool depicting spatio-temporal patterns of the forces' units over the course of a set of agent traces.

Temporal logical formulas provide an elegant formal mechanism for characterizing behavior over time; however, they must operate over semantically meaningful features to be translated into human-understandable descriptions. Such features can often be generated through straightforward translations of lower-level features, as we demonstrate with our feature extractor. As mentioned earlier, another option is to rely on the output of classifiers trained over the lower-level features. Currently, we are developing such mechanisms to learn high-level unit group ``activities'' that are not easily hand-coded and thus complement our extractor, \eg determining which unit group is firing at which opponent group.

Our approach uses clustering to partition the input set of traces and then performs strategy inference over each cluster. One potential limitation of this approach is that clustering is aimed at finding frequent patterns over the entire trajectory, whereas strategy inference aims to discover behavioral sequences corresponding to goal-directed agent behavior. Thus, the clusters may not always partition the traces in a way that would lead to distinct strategies being extracted for each of them. An alternative we have begun exploring is to apply strategy inference over the entire set of traces, thus focusing on the discovery of smaller ``tactics'' while letting higher-order strategies, composed of these tactics, to emerge.

In strategic tasks that require the achievement of sub-objectives toward the primary goal, there are often key decision points where the agent's decision has a significant impact on the outcome. We can think of the variety of paths through these decision points as the different strategies the agent can pursue to achieve a task. We could then find the different strategies by applying our existing approach to segments of traces bounded by these decision points. Our approach to strategy inference currently uses traces from a random agent to provide the baseline against which to compare the agent traces and find the formulas with maximal KL divergence scores. In this setting, there may be no corresponding subtraces in the random traces, since the random agent may not arrive at the same decision points. A challenge will thus be to find an appropriate set of baseline traces.

% ACKNOWLEDGMENTS
% =========================================
\section*{Acknowledgements}
This material is based upon work supported by the Defense Advanced Research Projects Agency (DARPA) under Contract No. HR001119C0112. Any opinions, findings and conclusions or recommendations expressed in this material are those of the author(s) and do not necessarily reflect the views of the DARPA.

% BIBLIOGRAPHY
% =========================================
\bibliographystyle{ACM-Reference-Format}
\bibliography{22-tbd-caml-y2}

% \newpage

% % APPENDIX
% % =========================================
% \appendix
% \section{Appendix if needed}%
% \label{Sec:Appendix}

% =========================================
\end{document}